\documentclass[10pt,twocolumn,letterpaper]{article}

\usepackage{wacv}              

\usepackage[accsupp]{axessibility} 

\usepackage{graphicx}
\usepackage{amsmath}
\usepackage{amssymb}
\usepackage{booktabs}
\usepackage{multirow}
\usepackage{placeins}
\usepackage{longtable}

\usepackage[pagebackref,breaklinks,colorlinks]{hyperref}

\usepackage[capitalize]{cleveref}
\crefname{section}{Sec.}{Secs.}
\Crefname{section}{Section}{Sections}
\Crefname{table}{Table}{Tables}
\crefname{table}{Tab.}{Tabs.}

\def\wacvPaperID{950} 
\def\confName{WACV}
\def\confYear{2024}

\newcommand\blfootnote[1]{%
  \begingroup
  \renewcommand\thefootnote{}\footnote{#1}%
  \addtocounter{footnote}{-1}%
  \endgroup
}

\begin{document}

\title{SwinIA: Self-Supervised Blind-Spot Image Denoising without Convolutions}

\vspace{-3mm}
\author{
\noindent
\begin{minipage}[t]{0.3\textwidth}
    \centering
    Mikhail Papkov*$^{1}$\\
    {\tt\small mikhail.papkov@ut.ee}
\end{minipage}
\hfill
\begin{minipage}[t]{0.3\textwidth}
    \centering
    Pavel Chizhov*$^{1,2}$\\
    {\tt\small pavel.chizhov@thws.de}
\end{minipage}
\hfill
\begin{minipage}[t]{0.3\textwidth}
    \centering
    Leopold Parts$^{1}$\\
    {\tt\small leopold.parts@ut.ee}
\end{minipage}
\vspace{3mm} \\
$^{1}$Institute of Computer Science, University of Tartu, Estonia \\
$^{2}$CAIRO, Technical University of Applied Sciences Würzburg-Schweinfurt, Germany
}

\maketitle

\begin{abstract}
    Self-supervised\blfootnote{*equal contribution} image denoising implies restoring the signal from a noisy image without access to the ground truth. State-of-the-art solutions for this task rely on predicting masked pixels with a fully-convolutional neural network. This most often requires multiple forward passes, information about the noise model, or intricate regularization functions. In this paper, we propose a Swin Transformer-based Image Autoencoder (SwinIA), the first fully-transformer architecture for self-supervised denoising. The flexibility of the attention mechanism helps to fulfill the blind-spot property that convolutional counterparts normally approximate. SwinIA can be trained end-to-end with a simple mean squared error loss without masking and does not require any prior knowledge about clean data or noise distribution. Simple to use, SwinIA establishes the state of the art on several common benchmarks.
\end{abstract}    
\section{Introduction}
\label{sec:intro}

Image denoising methods aim to reconstruct true signal given corrupted input. The corruption depends on the camera sensor, signal processor, and other aspects of the image acquisition procedure, and can take various forms such as Gaussian noise, Poisson noise, salt-and-pepper noise, \etc. Noise levels also vary with illumination and exposure, and some amount is always present in any image. This makes denoising an integral part of image processing pipelines. 

As in other fields in computer vision, deep learning solutions have superseded classical methods~\cite{dabov2007color,anscombe,buades2005non} for denoising. However, when approached naively, neural networks require huge amounts of paired noisy and clean images for supervised learning. Collecting such a dataset is usually impracticable. Lehtinen~\etal~\cite{lehtinen2018noise2noise} proposed Noise2Noise showing that supervision with independently corrupted data is equivalent to supervision with clean data. However, this approach still requires collecting multiple image copies, which may not be present in existing datasets. 

\begin{figure}[t]
\centering
\begin{subfigure}[t]{0.45\linewidth}
\begin{center}
    \includegraphics[width=0.6\linewidth]{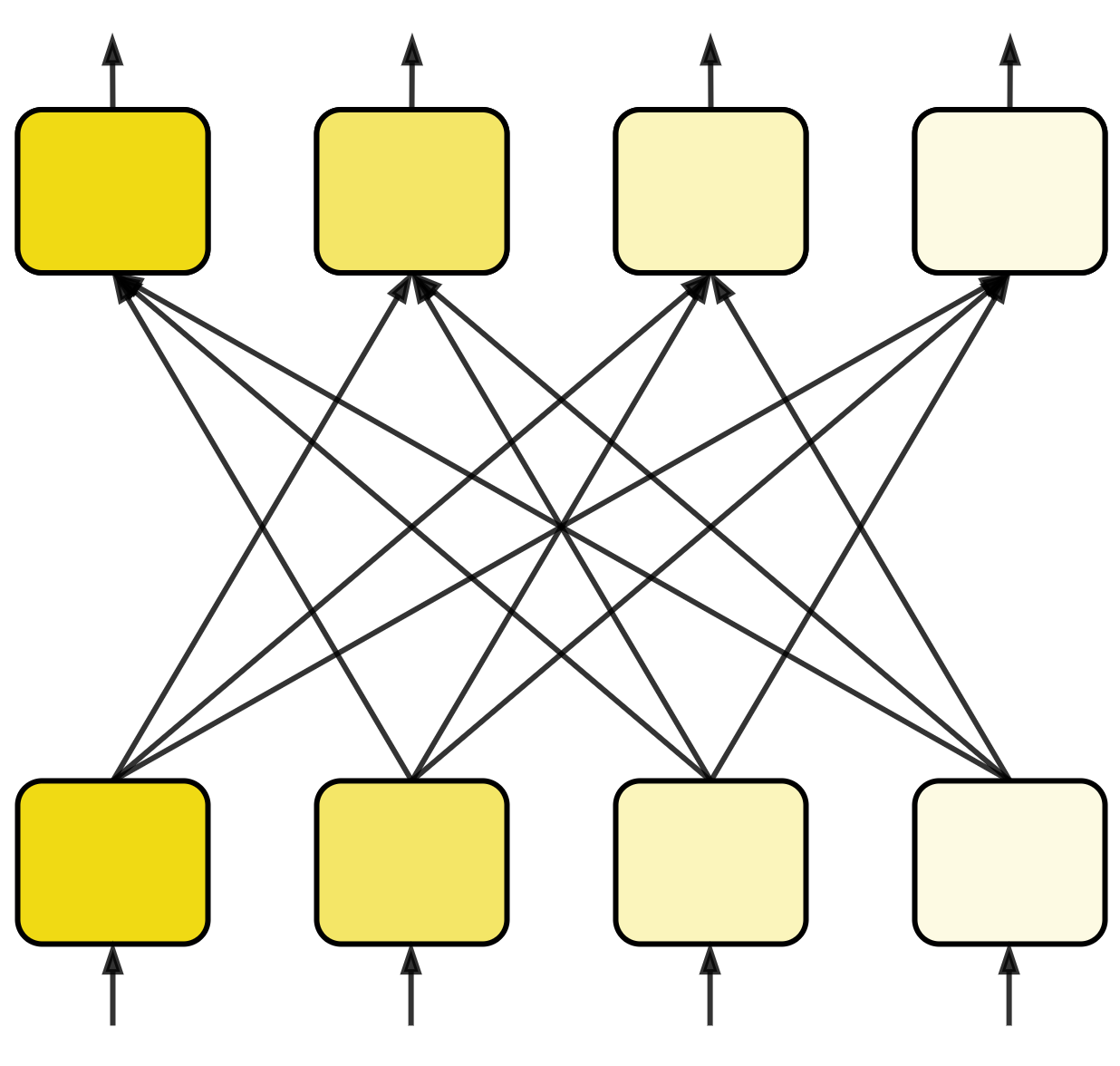}
   \caption{Each token representation is based on the rest of the sequence, excluding the token itself.}
    \label{fig:text-ae}
\end{center}
\end{subfigure}
\hfill
\begin{subfigure}[t]{0.45\linewidth}
\begin{center}
    \includegraphics[width=0.7\linewidth]{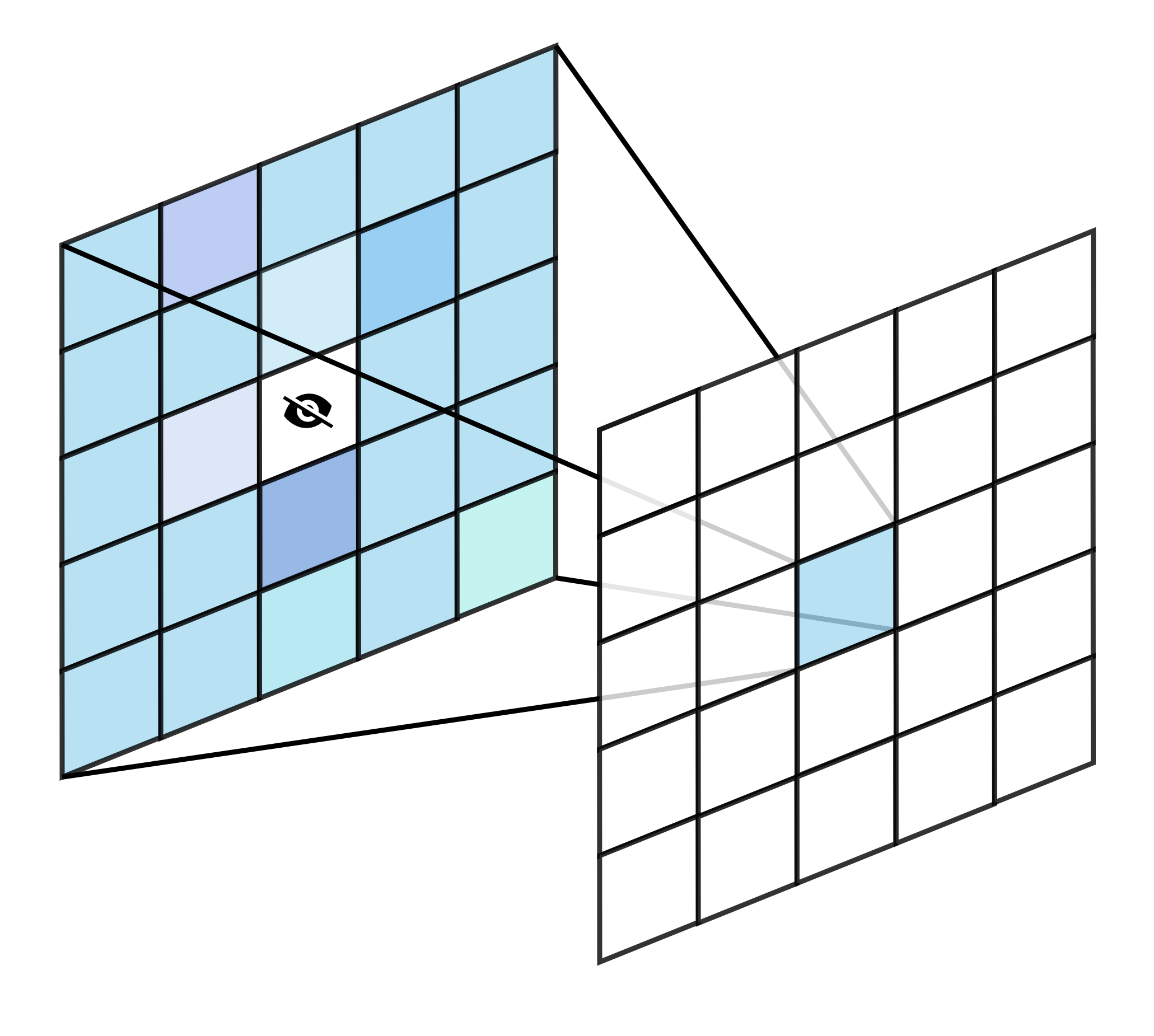}
    \caption{Each pixel representation is based on all other pixels, making its own value a blind spot.}
    \label{fig:image-ae}
\end{center}
\end{subfigure}
    \caption{Self-unaware autoencoding in text and images.}
    \label{fig:ae}
\end{figure}

Self-supervised denoising avoids the demand for paired data since it learns only from noisy images. Developing Noise2Noise ideas, Noisier2Noise~\cite{moran2020noisier2noise} and Recorrupted2Recorrupted~\cite{pang2021recorrupted} applied additional noise on training images to emulate the strongly supervised Noise2Noise scenario. This can be accomplished by assuming a prior knowledge of the noise model or test-time aggregation. Self2Self~\cite{quan2020self2self} proposed training a network with Bernoulli input dropout and inference by ensembling multiple outputs. Neighbor2Neighbor~\cite{huang2021neighbor2neighbor} sub-sampled the input image and treated the result as independently corrupted copies. 

Noise2Void~\cite{n2void} and Noise2Self~\cite{n2self} introduced a different approach to self-supervised denoising~--- a blind-spot network (BSN). This type of network reconstructs a pixel from its neighborhood, assuming spatially independent zero-mean noise. It is practically difficult to dissect the continuous receptive field of convolutional neural networks (CNN), so BSN is usually emulated by a masking procedure to hide a small portion of pixels by substitution or random noise and learn solely from them. However, learning from a few data points per image slows down convergence, and different masking approaches may produce drastically different results~\cite{n2same}. Laine~\etal~\cite{laine2019high} proposed to restrict blindness by constructing four denoising branches with unidirectional receptive fields. In practice, this was achieved by passing four differently rotated input copies through the network. Later, Honz{\'a}tko~\etal~\cite{honzatko2020efficient} and Wu~\etal~\cite{wu2020unpaired} adopted dilated convolutions to create a true BSN which does not require masking. We further discuss these methods below. Subsequent works abandoned the idea of strict pixel blindness and adopted multiple forward passes through the network. Noise2Same~\cite{n2same} and its modification Noise2Info~\cite{n2info} make two forward passes (one with a random mask, one without) and regularize the training with invariance loss in masked pixel locations. DCD-Net~\cite{Zou_2023_ICCV} combines Noise2Noise and Noise2Void in an iterative denoise-corrupt-denoise pipeline. Blind2Unblind~\cite{wang2022blind2unblind} utilizes global masking and combines the denoising results from passing 17 
image copies\footnote{In the official repository, authors set mask window width to 4, creating 16 masked copies in addition to the unmasked image.} through the network. While superior in performance, this approach is time-consuming, requires tuning multiple hyperparameters for each dataset, and exhibits unstable training (\cref{tab:imagenet-hanzi}).

Most recently, vision transformers started to outperform CNNs across a variety of benchmarks, including supervised denoising. SwinIR~\cite{swinir}, based on Swin Transformer~\cite{swin}, achieved state-of-the-art results in JPEG compression artifact reduction. Uformer~\cite{wang2022uformer} and Restormer~\cite{zamir2022restormer} concurrently excelled in camera noise removal. Evolution of vision transformers closely followed their path in natural language processing~\cite{dosovitskiy2020image}. Bao~\etal~\cite{bao2021beit} introduced BERT-style pre-training for image datasets, and He~\etal~\cite{he2022masked} showed that transformers can confidently reconstruct up to 95\% of hidden data in a masked autoencoder fashion. These results hint that we could use transformers to design blind-spot self-supervised denoising models.

In this paper, we propose \textbf{SwinIA} --- Swin Transformer-based Image Autoencoder, the first fully transformer-based architecture for self-supervised image denoising. SwinIA does not require any prior knowledge of noise distribution. It also does not have access to clean images, either through pre-training or knowledge distillation. Neither does it use input masking, auxiliary regularization losses, or multiple forward passes. Instead, SwinIA is trained as a plain autoencoder by minimizing the mean squared error (MSE) computed over the full image. To our knowledge, it is the first precise implementation of the original BSN idea.
We rigorously test our SwinIA method on a variety of synthetic and real-world datasets and demonstrate its competitiveness against state-of-the-art self-supervised denoising solutions.
\section{Related work}
\label{sec:related}
We further describe the foundational ideas of blind-spot networks and denoising vision transformers that our model is related to. We introduce their properties and usual training schemes and give the context for our advances. 

The blind-spot property is usually achieved by masking~\cite{n2void} or multiple forward passes through the network~\cite{laine2019high}. These techniques overcome the continuous receptive field issue of CNNs. It is also possible to maintain blindness with increasingly dilating convolutions. Honz{\'a}tko~\etal~\cite{honzatko2020efficient} proposed a blind-spot convolutional layer with a virtual ``hole'' in the kernel center. Their work followed the training setup of Laine~\etal~\cite{laine2019high} and demonstrated similar performance on sRGB datasets. The main limitation of both approaches is the assumption that the noise distribution is known and the predictions are refined with probabilistic post-processing (posterior mean estimation, or PME). Finally, Wu~\etal~\cite{wu2020unpaired} also utilized a dilated blind-spot network (DBSN) in a multi-stage pipeline with clean images provided via knowledge distillation.

Transformers are widely used for image restoration in the supervised setting~\cite{swinir,wang2022uformer,zamir2022restormer}, but rarely for self-supervised denoising. Zhang~\etal~\cite{zhang2023self} proposed a Context-aware Denoise Transformer (CADT) based on SwinIR~\cite{swinir} and masking scheme of Blind2Unblind~\cite{wang2022blind2unblind}. They used Swin Transformer~\cite{swin} blocks in the global branch of the network and trained them with patch embeddings, not pixel embeddings. However, they argued that a transformer alone is not suitable for the task and thus complemented it with convolutional local feature extraction. CADT used sixteen masked forward passes from Blind2Unblind in both training and inference. Liu~\etal~\cite{dnt} built a single-image denoising transformer (DnT) from self-attention blocks interleaved with convolutional layers. This architecture was not tested for self-supervised denoising on larger datasets.
\section{Design}
\label{sec:design}

BSN was proposed many years ago~\cite{n2void,n2self}, but to date, there is no implementation strictly adhering to the original idea. Existing solutions use masking~\cite{n2void,n2self,lee2022ap,wang2022blind2unblind}, assume known noise distributions~\cite{laine2019high,honzatko2020efficient}, or learn from clean data through knowledge distillation~\cite{wu2020unpaired}. Thus, creating an assumption-free BSN that is trained end-to-end as an autoencoder with hyperparameter-free MSE loss between input and output remains an open challenge. This learning objective can be formulated as follows:
\begin{equation}
\label{eq:mse}
\mathcal{L}(f|\theta)=\mathbb{E}_x\lVert f(x|\theta) - x \rVert^2.
\end{equation}
Here $x$ is a noisy input image and $f$ is a model with a set of parameters $\theta$. We hypothesize that transformers could be suitable for this task because it is possible to control pixel interaction through the attention mechanism.

\begin{figure*}[t!]
\centering
\includegraphics[width=\linewidth]{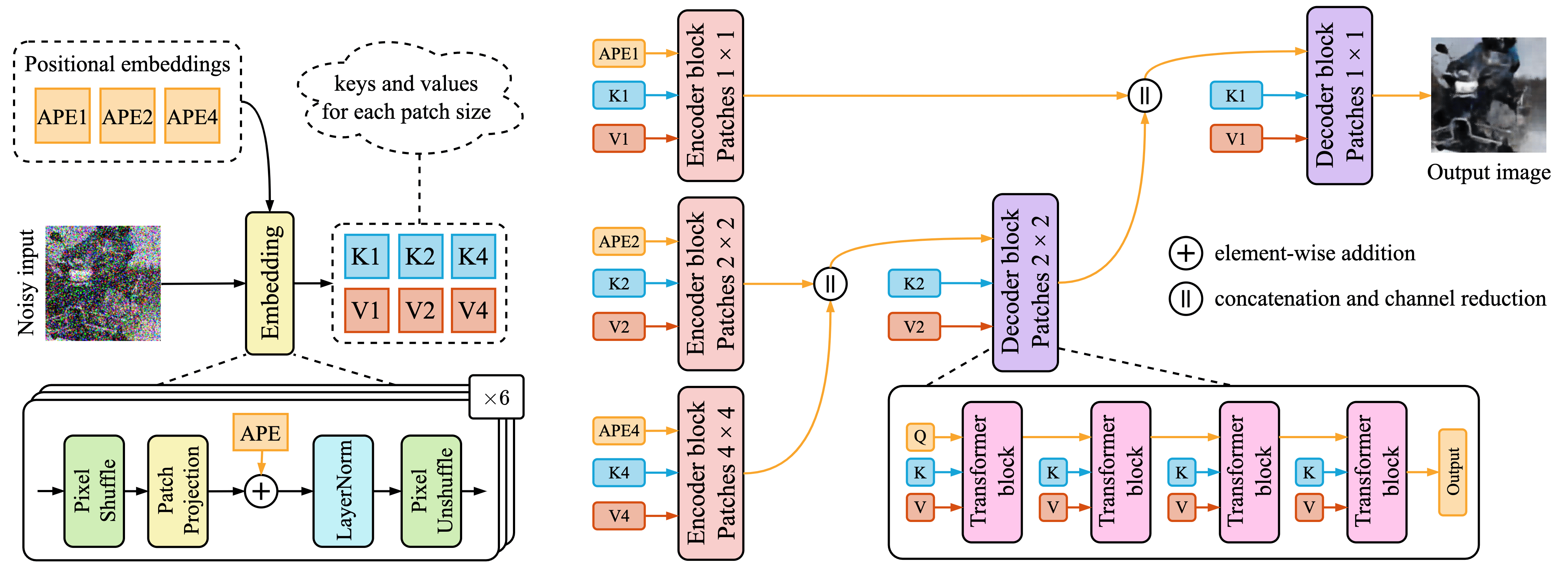}
\caption{SwinIA model. Multiscale positional embeddings act as queries for the encoder (three parallel blocks) and are added to patch embeddings to create constant keys and values. The decoder (two remaining blocks) fuses the extracted features into the final output image. Encoder and decoder blocks have an identical structure and consist of four transformer blocks with cyclically shifted window attention.}
\label{fig:swinia}
\end{figure*}

Shin~\etal~\cite{shin-etal-2020-fast} introduced the idea of self-unaware text autoencoding using transformers (T-TA). They modified the transformer model so that each token representation is built based on all the other tokens, except itself, as in \cref{fig:text-ae}. T-TA builds text representations in one iteration without access to a token's own value, as opposed to the masked language modeling objective of BERT~\cite{devlin-etal-2019-bert}, where the tokens are processed one at a time. We transfer this idea to the image domain to create a vision transformer autoencoder with self-unaware pixel-level tokens (\cref{fig:image-ae}).

In order to efficiently exploit a transformer-based model at the pixel level for high-dimensional image data, we need to use local attention. Swin Transformer~\cite{swin} is a powerful multi-purpose vision model, which uses the window multi-head self-attention (MSA) mechanism that restricts self-attention to windows of fixed size. The windows are shifted from block to block to avoid bordering artifacts and spread the field of view. Swin Transformer was already efficiently utilized at the pixel level in SwinIR~\cite{swinir}, where individual pixels were embedded into tokens.

Combining the ideas above, we formulate a list of requirements to design a blind-spot transformer.

\textbf{Pixel level}. The network should process images at single-pixel level. The absence of pixel processing would impede the understanding of random pixel-level noise and the reconstruction of high-frequency details.

\textbf{Self-unawareness}. At any stage of the blind-spot network, individual pixels should not have access to their own state on the previous levels. This will prevent it from learning an identity function by minimizing MSE loss.

\textbf{Unblinding}. Blind-spot training inevitably leads to information loss from the most significant source~--- the actual value of the pixel. Therefore, it is important to unhide these values during inference without disturbing the learned modality of the model. 

\textbf{Long-range interactions}. 
In our setting, downsampling pooling operations in the encoder would disrupt input isolation by mixing together feature vectors of individual pixels. Therefore, we need a downsampling operation that enables attention between groups of pixels and at the same time, maintains the independence of each pixel.

\section{Methods}
\label{sec:methods}

\subsection{Input embedding}
\begin{figure}[t]
\centering
\includegraphics[width=0.9\linewidth]{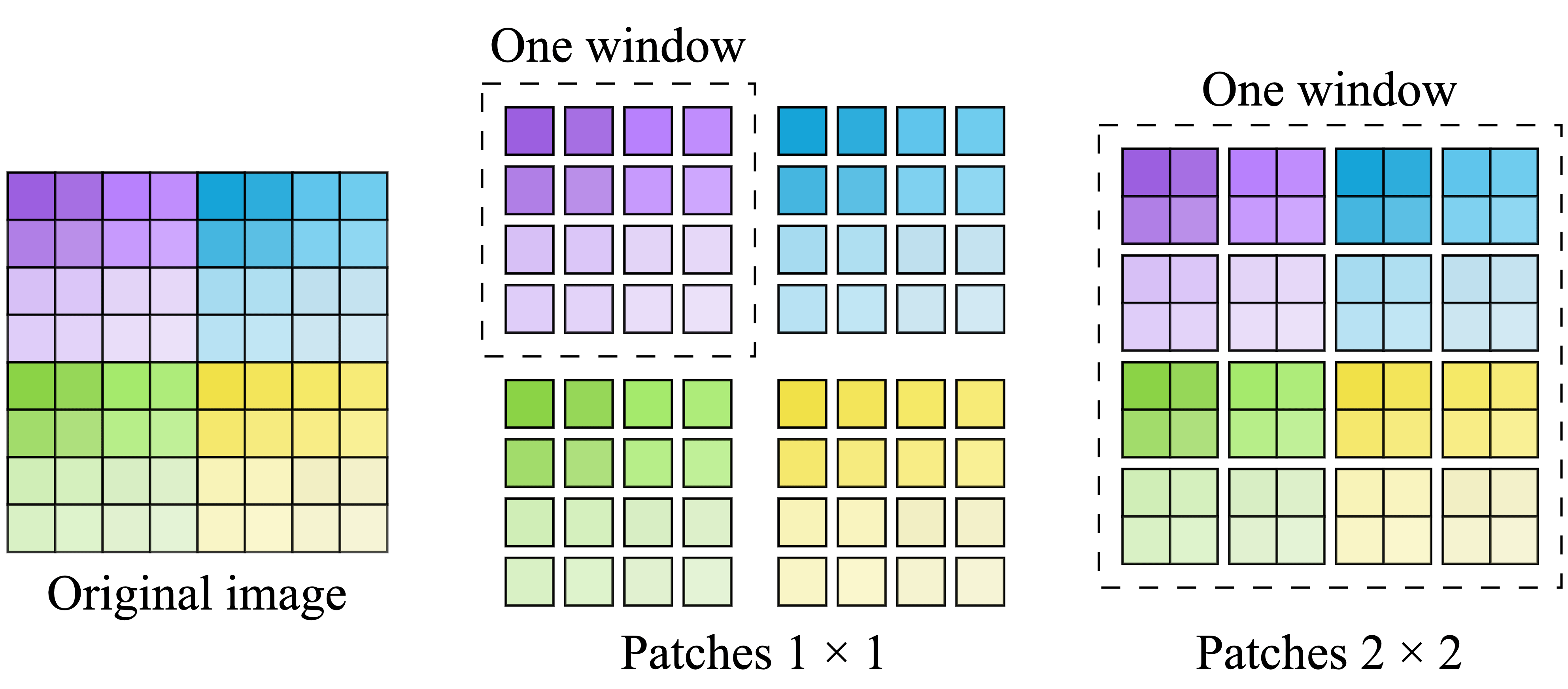}
\caption{Pixel shuffle example of an image of size $8\times 8$ with window size~$4$ into patches of sizes $1\times1$ and $2\times2$.}
\label{fig:shuffle}
\end{figure}

In contrast with the explicit pixel-level processing in SwinIR~\cite{swinir}, we propose to operate on shuffled square patches of size $p\times p$ pixels (see the left part of~\cref{fig:swinia}). An example of pixel shuffle is illustrated in \cref{fig:shuffle}. The {queries} are set to learnable absolute positional embeddings (APE), separate for each patch size. The input image is projected into {keys} and {values} only once for each patch size $p\in\left\{1,2,4\right\}$ as follows:
\begin{align}
\label{eq:kv}
K_p = h_p^{-1}\left(\mathrm{LayerNorm}\left(h_p\left(X\right)W_{kp} + \mathrm{APE}_p\right)\right), \\
V_p = h_p^{-1}\left(\mathrm{LayerNorm}\left(h_p\left(X\right)W_{vp} + \mathrm{APE}_p\right)\right).
\end{align}
Here $W_{kp}, W_{vp}$ are linear projection parameter matrices, and $h_p$ is an operation of shuffling into patches of size $p\times p$. Maintaining keys and values intact throughout the architecture is essential for self-unawareness~\cite{shin-etal-2020-fast}.

\subsection{SwinIA model}

SwinIA is an encoder-decoder model consisting of three encoder and two decoder blocks, the architecture is presented in the left part of \cref{fig:swinia}. Three encoder blocks encode the inputs for each of the three patch sizes $p\in\left\{1,2,4\right\}$ and are computed separately, each using a corresponding set of positional embeddings (APE) as queries. The encoded representations are fused up through the decoders of corresponding patch sizes with skip connections by concatenation and linear projection:
\begin{equation}
\label{eq:shortcut}
\mathrm{shortcut(X_1, X_2)} = \left(X_1 \mathbin\Vert X_2\right) W + b.
\end{equation}

Each encoder/decoder block consists of four transformer blocks. As in Swin Transformer~\cite{swin}, the attention is computed in square windows of fixed size. To avoid bordering artifacts, the images have to be diagonally shifted in every second block. The shift size along one dimension is equal to half of a window size.

\subsection{Transformer block}
\label{sec:transformer}

\begin{figure}[t]
\centering
\includegraphics[width=0.9\linewidth]{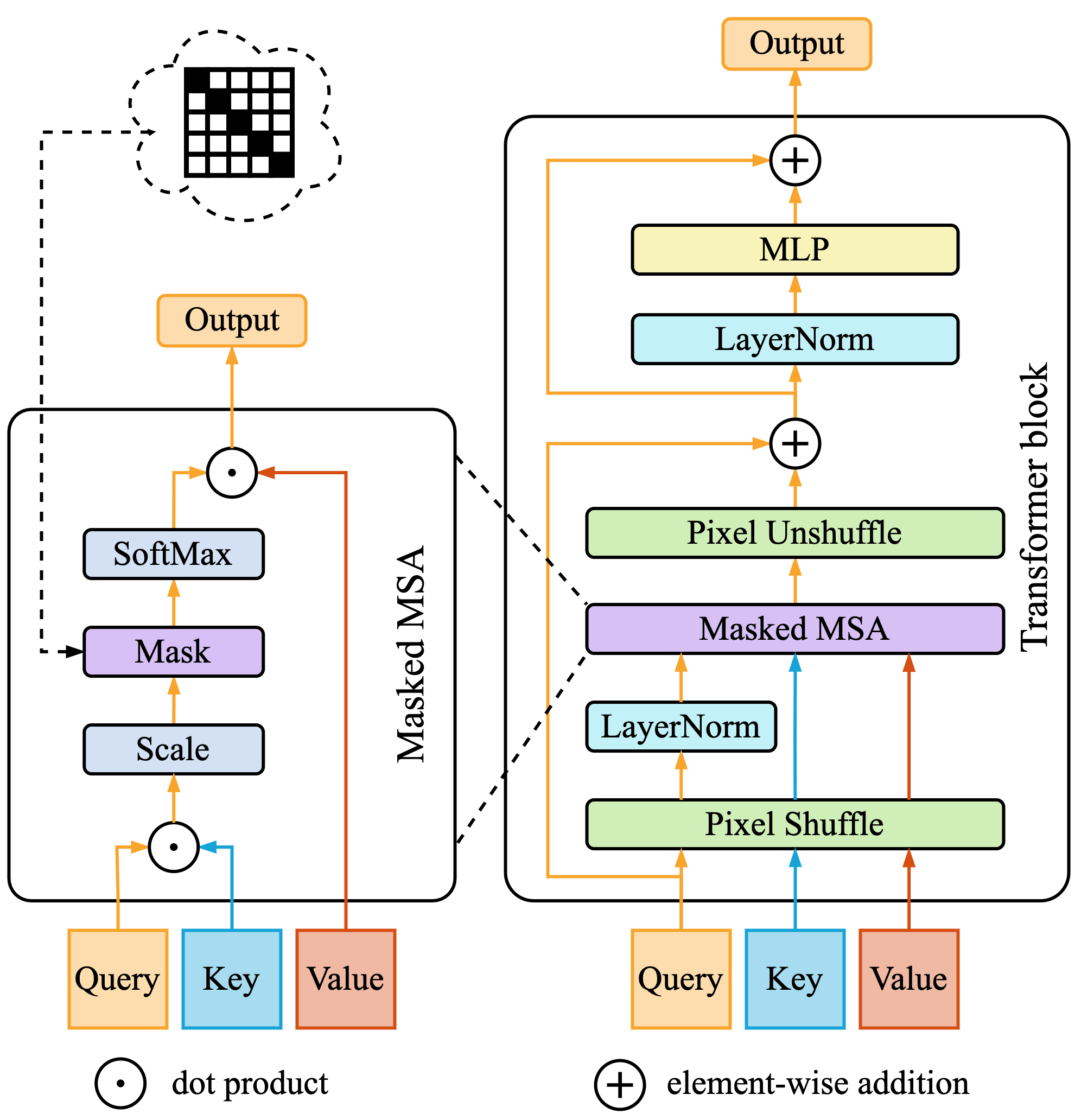}
\caption{SwinIA transformer block architecture. MSA and MLP are preceded by layer normalization and complemented with a shortcut by addition. Only queries are normalized before the MSA because keys and values are normalized upon creation. The attention is performed between shuffled patches. The attention matrix is diagonally masked to maintain pixel self-unawareness.}
\label{fig:trans}
\end{figure}

Transformer blocks in SwinIA are comprised of multi-head self-attention (MSA) and multi-layer perceptron (MLP) submodules with pre-normalization and additive shortcuts, as in~\cref{fig:trans}. Since we compute patch-level attention, the inputs are first shuffled into patches as 2D matrices $\mathbb{R}^{p^2\times d}$, where $p$ is patch size along one dimension and $d$ is the embedding dimensionality in the model.

SwinIA transformer block utilizes a window multi-head self-attention (MSA) mechanism with a masked main diagonal of the attention matrix. The masking is performed by subtracting a large constant from the main diagonal of the dot-product, therefore the SoftMax values there become infinitesimal:
\begin{equation}
\label{eq:msa}
\mathrm{MSA}\left(Q,K,V\right) = \mathrm{SoftMax}\left(\frac{QK^T-I_n\cdot10^{9}}{\sqrt{d_h}\cdot p}\right)V.
\end{equation}
Here $d_h$ is embedding dimensionality per attention head, $p$ is the current patch size, and $I_n$ is an identity matrix where $n$ is the attention sequence length.

The MSA is followed by a layer normalization and pixel unshuffle operation. The unshuffled outputs are fed into a two-layer MLP with 4 times increased hidden dimensionality and GELU activations~\cite{Hendrycks2016}. 

\subsection{Architecture justification}

In this part, we will analyze how the proposed architecture addresses the design requirements formulated in \cref{sec:design}.

SwinIA operates on a \textbf{pixel level}, because its smallest patch size is $1\times1$. Therefore, our model can capture valuable pixel-to-pixel interactions.

\textbf{Self-unawareness} in SwinIA is ensured by a combination of the diagonal attention mask and input isolation. The diagonal mask restricts the attention so that none of the pixels have access to their value from the previous layer. However, this restriction could be bypassed by a simple permutation learning within two consequent transformer blocks. Input isolation makes it impossible: in every dot-product of the attention, only one of the components is aware of its surroundings, as keys and values are projected with a single-patch field of view and frozen from the beginning.

Additionally, unlike in standard encoder-decoder architectures, encoder blocks in SwinIA run in parallel. Since the patch size increases in the encoder flow, bigger patches would consist of context-aware smaller patches from the previous level. As a result, the noise would leak, and the model would learn a simple identity function (see \cref{tab:ablation}).

\begin{table*}[t!]
\small
  \centering
  \begin{tabular}[b]{clcccccccc}
    \toprule
      & & \multirow{2}{*}{\rotatebox{90}{train}} & \multirow{2}{*}{\rotatebox{90}{test}} & \multicolumn{2}{c}{Gaussian $\sigma=15$} & \multicolumn{2}{c}{Gaussian $\sigma=25$} & \multicolumn{2}{c}{Gaussian $\sigma=50$} \\
      & Method &  &  & BSD68 & Set12 & BSD68 & Set12 & BSD68 & Set12 \\
    \midrule
      \multirow{2}{*}{\textit{Supervised}} & Noise2Clean & 1 & 1 & 31.58/0.889 & 32.60/0.899 & 29.02/0.822 & 30.07/0.852 & 26.08/0.715 & 26.88/0.777 \\
      & SwinIR & 1 & 1 & 31.97/\hphantom{0}---\hphantom{.0} & 33.36/\hphantom{0}---\hphantom{.0} & 29.50/\hphantom{0}---\hphantom{.0} & 31.01/\hphantom{0}---\hphantom{.0} & 26.58/\hphantom{0}---\hphantom{.0} & 27.91/\hphantom{0}---\hphantom{.0} \\
    \midrule
      \multirow{4}{*}{\textit{Self-sup.}} & R2R & 1 & 50 & \underline{31.54}/\underline{0.885} & \textbf{32.54}/\textbf{0.897} & 28.99/0.818 & \underline{30.06}/\underline{0.851} & 26.02/0.705 & 26.86/0.771 \\
      & Noise2Self\textsuperscript{$\dag$} & 1 & 1 & 30.63/0.843 & 29.88/0.840 & 28.88/0.789 & 28.37/0.799 & 26.19/0.664 & 25.56/0.692 \\
      & Noise2Same\textsuperscript{$\dag$} & 2 & 1 & 30.85/0.850 & 30.02/0.849 & \underline{29.13}/0.800 & 28.54/0.814 & \underline{26.75}/0.714 & 26.13/0.744 \\
      & Blind2Unblind & 17 & 1 & 31.44/0.884 & \underline{32.46}/\textbf{0.897} & 28.99/\underline{0.820} & \textbf{30.09}/\textbf{0.854} & 26.09/\underline{0.715} & \textbf{26.91}/\textbf{0.776} \\
    \midrule
      \multirow{2}{*}{\shortstack[c]{\textit{Self-sup.} \\ \textit{(true BSN)}}} & Laine19 & 4 & 4 & --- & --- & 28.84/0.814 & --- & 25.78/0.698 & --- \\
      & SwinIA (ours)\textsuperscript{$\dag$} & 1 & 1 & \textbf{31.84}/\textbf{0.885} & 31.04/0.882 & \textbf{30.01}/\textbf{0.837} & 29.61/0.848 & \textbf{27.23}/\textbf{0.743} & \underline{26.88}/\underline{0.772} \\
    \bottomrule
  \end{tabular}
  \caption{Grayscale image denoising results for BSD68 and Set12 with synthetic noise along with the method description (supervision type and number of train and test passes). The highest PSNR(dB)/SSIM among self-supervised denoising methods is highlighted in \textbf{bold}, the second-best is \underline{underlined}. \textsuperscript{$\dag$} denotes the models that we implemented and trained ourselves.}
  \label{tab:gray}
\end{table*}

\textbf{Unblinding} is achieved during inference by removing the diagonal mask and thus applying the complete set of attention weights to values:
\begin{equation}
\label{eq:msai}
\mathrm{MSA}_\mathrm{eval}\left(Q,K,V\right) = \mathrm{SoftMax}\left(\frac{QK^T}{\sqrt{d_h}\cdot p}\right)V.
\end{equation}
Intuitively, this allows to propagate pixel's own signal by iteratively re-weighting it with the most similar neighbors throughout the network. Since attention matrix does not contain learnable components and simply reflects (self\nobreakdash-)similarity of pixel embeddings, unmasking the main diagonal maintains the learned modality and does not disrupt the forward pass. 

\section{Experimental results}
\label{sec:results}

For our experiments, we chose a model configuration with embeddings of dimensionality $144$ throughout the network, $16$ attention heads in each block, and windows of size $8\times 8$. We extensively test SwinIA against state-of-the-art self-supervised denoising methods on synthetic and real-world data. Since SwinIA is a BSN and inevitably loses information because of hiding pixels from themselves, we separately focus on comparison with methods with similar properties~\cite{laine2019high}. We use peak signal-to-noise ratio (PSNR) and structural similarity index (SSIM) for evaluation. Training process is described in detail in~\cref{sec:training}. 

\subsection{Synthetic noise (grayscale)}
\label{sec:synth-gray}
Following~Wang~\etal~\cite{wang2022blind2unblind}, we use BSD400~\cite{zhang2017beyond} for training and test on Set12 and BSD68~\cite{roth2005fields}.We apply Gaussian noise with $\sigma=\{15,25,50\}$ to the images. For evaluation, we repeat BSD68 4 times, and Set12 20 times. This results in 512 (272 and 240) testing images in total. 

The results are summarized in \cref{tab:gray}. On BSD68, SwinIA ranked first for all noise levels. Apart from self-supervised methods, SwinIA beats SwinIR --- a supervised denoising transformer. Interestingly, we obtained lower scores on Set12 but still consistently outperformed Noise2Self and Noise2Same and tailed the scores of R2R and Blind2Unblind. We also improved over the other BSN~\cite{laine2019high} by $+1.31$dB PSNR on BSD68 on average.

\begin{table}[t!]
\small
  \centering
  \begin{tabular}[b]{lcccc}
    \toprule
     Method & train & test & ImageNet & H\`anZ\`i \\
     \midrule
     NLM & - & - & 18.04/\hphantom{0}---\hphantom{.0} & \hphantom{0}8.41/\hphantom{0}---\hphantom{.0} \\
     BM3D & - & - & 18.74/\hphantom{0}---\hphantom{.0} & 10.90/\hphantom{0}---\hphantom{.0} \\
     \midrule
     Noise2Clean & 1 & 1 & 23.39/\hphantom{0}---\hphantom{.0} & 15.66/\hphantom{0}---\hphantom{.0}  \\
     Noise2Noise & 1 & 1 & 23.27/\hphantom{0}---\hphantom{.0} & 14.30/\hphantom{0}---\hphantom{.0} \\ \midrule
     Noise2Void & 1 & 1 & 21.36/\hphantom{0}---\hphantom{.0} & 13.72/\hphantom{0}---\hphantom{.0} \\
     Noise2Self\textsuperscript{$\dag$} & 1 & 1 & 21.33/0.574 & 14.16/0.512 \\
     Noise2Same\textsuperscript{$\dag$} & 2 & 1 & 22.85/0.625 & \underline{14.85}/\underline{0.542} \\ 
     Blind2Unblind$^{\ast}$  & 17 & 1 & \underline{23.74}/\underline{0.649} & 13.87/0.509 \\ 
     Noise2Info & 2 & 1 & 22.60/\hphantom{0}---\hphantom{.0} & 14.43/\hphantom{0}---\hphantom{.0} \\
     \midrule 
     Laine19 & 4 & 4 & 20.89/\hphantom{0}---\hphantom{.0} & 10.70/\hphantom{0}---\hphantom{.0} \\
     SwinIA (ours)\textsuperscript{$\dag$}  & 1 & 1 & \textbf{23.91}/\textbf{0.668} & \textbf{14.92}/\textbf{0.574} \\  
    \bottomrule
  \end{tabular}
  \caption{Denoising results on datasets with mixed synthetic noise along with the method description (number of train and test passes). The highest PSNR(dB)/SSIM among self-supervised denoising methods is in \textbf{bold}, while the second-best is \underline{underlined}. \textsuperscript{$\dag$} denotes the models that we implemented and trained ourselves. $^{\ast}$Blind2Unblind diverged on H\`anZ\`i with different learning rates, so we provide average metrics of three runs after the 20th epoch.}
  \label{tab:imagenet-hanzi}
\end{table}

\begin{table*}[t] 
\small
  \centering
  \begin{tabular}{clcccccccc}
    \toprule
      &&\multirow{2}{*}{\rotatebox{90}{train}} & \multirow{2}{*}{\rotatebox{90}{test}} & \multicolumn{3}{c}{Gaussian $\sigma=25$} & \multicolumn{3}{c}{Gaussian $\sigma \in \left[5, 50\right]$} \\
      & Method & &  & KODAK & BSD300 & SET14 & KODAK & BSD300 & SET14 \\
    \cmidrule(r){1-4} \cmidrule(lr){5-7} \cmidrule(lr){8-10}
      \textit{Traditional} & CBM3D & - & - & 31.87/0.868 & 30.48/0.861 & 30.88/0.854 & 32.02/0.860 & 30.56/0.847 & 30.94/0.849 \\ 
    \cmidrule(r){1-4} \cmidrule(lr){5-7} \cmidrule(lr){8-10}
      \multirow{2}{*}{\textit{Supervised}} & Noise2Clean & 1 & 1 & 32.43/0.884 & 31.05/0.879 & 31.40/0.869 & 32.51/0.875 & 31.07/0.866 & 31.41/0.863 \\
      & Noise2Noise & 1 & 1 & 32.41/0.884 & 31.04/0.878 & 31.37/0.868 & 32.50/0.875 & 31.07/0.866 & 31.39/0.863 \\
    \cmidrule(r){1-4} \cmidrule(lr){5-7} \cmidrule(lr){8-10}
      \multirow{3}{*}{\shortstack[c]{\textit{Self-sup.} \\ \textit{(Gaussian)$^{\ast}$}}} &Laine19-pme & 4 & 4 & 32.40/0.883 & 30.99/0.877 & 31.36/0.866 & 32.40/0.870 & 30.95/0.861 & 31.21/0.855 \\
      &Honzatko-pme & 1 & 1 & 32.45/\hphantom{0}---\hphantom{.0} & 31.02/\hphantom{0}---\hphantom{.0} & 31.25/\hphantom{0}---\hphantom{.0} & 32.46/\hphantom{0}---\hphantom{.0} & 31.18/\hphantom{0}---\hphantom{.0} & 31.25/\hphantom{0}---\hphantom{.0} \\
      &Noisier2Noise & 1 & 1 & 30.70/0.845 & 29.32/0.833 & 29.64/0.832 & --- & --- & --- \\
    \cmidrule(r){1-4} \cmidrule(lr){5-7} \cmidrule(lr){8-10}
      \multirow{6}{*}{\textit{Self-sup.}} &Self2Self & 1 & 50 & 31.28/0.864 & 29.86/0.849 & 30.08/0.839 & 31.37/0.860 & 29.87/0.841 & 29.97/0.849 \\
      &Noise2Void & 1 & 1 & 30.32/0.821 & 29.34/0.824 & 28.84/0.802 & 30.44/0.806 & 29.31/0.801 & 29.01/0.792 \\
      &Noise2Same\textsuperscript{$\dag$} & 2 & 1 & 30.77/0.841 & 29.50/0.834 & 29.53/0.827 & 30.78/0.835 & 29.49/0.823 & 29.34/0.817 \\   
      &DBSN & 2 & 1 & 31.64/0.856 & 29.80/0.839 & 30.63/0.846 & 30.38/0.826 & 28.34/0.788 & 29.49/0.814 \\
      &R2R & 1 & 50 & 32.25/\underline{0.880} & \underline{30.91}/0.872 & \textbf{31.32}/\textbf{0.865} & 31.50/0.850 & 30.56/0.855 & 30.84/0.850 \\
      &NBR2NBR & 2 & 1 & 32.08/0.879 & 30.79/\underline{0.873} & 31.09/\underline{0.864} & 32.10/0.870 & 30.73/\underline{0.861} & 31.05/\textbf{0.858} \\  
      &B2UB & 17 & 1 & \textbf{32.27}/\underline{0.880} & 30.87/0.872 & 31.27/\underline{0.864} & \underline{32.34}/\textbf{0.872} & \underline{30.86}/\underline{0.861} & \textbf{31.14}/\underline{0.857} \\
      &DCD-Net & 3 & 1 & \textbf{32.27}/\textbf{0.881} & \textbf{31.01}/\textbf{0.876} & \underline{31.29}/0.862 & \textbf{32.35}/\textbf{0.872} & \textbf{31.09}/\textbf{0.866} & \underline{31.09}/0.855 \\
    \cmidrule(r){1-4} \cmidrule(lr){5-7} \cmidrule(lr){8-10} 
      \multirow{2}{*}{\shortstack[c]{\textit{Self-sup.} \\ \textit{(true BSN)}}} & Laine19 & 4 & 4 & 30.62/0.840 & 28.62/0.803 & 29.93/0.830 & 30.52/0.833 & 28.43/0.794 & 29.71/0.822 \\
      &SwinIA (ours)\textsuperscript{$\dag$} & 1 & 1 & 31.43/0.863 & 29.94/0.853 & 30.56/0.856 & 31.54/0.859 & 30.00/0.847 & 30.55/0.849 \\
    \bottomrule
    \toprule
      &&\multirow{2}{*}{\rotatebox{90}{train}} & \multirow{2}{*}{\rotatebox{90}{test}} & \multicolumn{3}{c}{Poisson $\lambda=30$} & \multicolumn{3}{c}{Poisson $\lambda \in \left[5, 50\right]$} \\
      &Method &  &  & KODAK & BSD300 & SET14 & KODAK & BSD300 & SET14 \\
    \cmidrule(r){1-4} \cmidrule(lr){5-7} \cmidrule(lr){8-10}
      \textit{Traditional} & Anscombe & - & - & 30.53/0.856 & 29.18/0.842 & 29.44/0.837 & 29.40/0.836 & 28.22/0.815 & 28.51/0.817 \\ 
    \cmidrule(r){1-4} \cmidrule(lr){5-7} \cmidrule(lr){8-10}
      \multirow{2}{*}{\textit{Supervised}} & Noise2Clean & 1 & 1 & 31.78/0.876 & 30.36/0.868 & 30.57/0.858 & 31.19/0.861 & 29.79/0.848 & 30.02/0.842 \\ 
      &Noise2Noise & 1 & 1  & 31.77/0.876 & 30.35/0.868 & 30.56/0.857 & 31.18/0.861 & 29.78/0.848 & 30.02/0.842 \\
    \cmidrule(r){1-4} \cmidrule(lr){5-7} \cmidrule(lr){8-10}
      \multirow{2}{*}{\shortstack[c]{\textit{Self-sup.} \\ \textit{(Poisson)$^{\ast}$}}} &Laine19-pme & 4 & 4 & 31.67/0.874 & 30.25/0.866 & 30.47/0.855 & 30.88/0.850 & 29.57/0.841 & 28.65/0.785 \\
      &Honzatko-pme & 1 & 1 & 31.67/\hphantom{0}---\hphantom{.0}  & 30.25/\hphantom{0}---\hphantom{.0} & 30.14/\hphantom{0}---\hphantom{.0} & --- & --- & ---\\
    \cmidrule(r){1-4} \cmidrule(lr){5-7} \cmidrule(lr){8-10}
      \multirow{6}{*}{\textit{Self-sup.}} &Self2Self &  1 & 50  & 30.31/0.857 & 28.93/0.840 & 28.84/0.839 & 29.06/0.834 & 28.15/0.817 & 28.83/0.841 \\
      &Noise2Void & 1 & 1  & 28.90/0.788 & 28.46/0.798 & 27.73/0.774 & 28.78/0.758 & 27.92/0.766 & 27.43/0.745 \\
      &Noise2Same\textsuperscript{$\dag$} & 2 & 1 & 27.73/0.747 & 26.69/0.714 & 26.78/0.735 & 27.44/0.738 & 26.36/0.700 & 26.37/0.721 \\    
      &DBSN & 2 & 1  & 30.07/0.827 & 28.19/0.790 & 29.16/0.814 & 29.60/0.811 & 27.81/0.771 & 28.72/0.800 \\
      &R2R & 1 & 50 & 30.50/0.801 & 29.47/0.811 & 29.53/0.801 & 29.14/0.732 & 28.68/0.771 & 28.77/0.765 \\
      &NBR2NBR & 2 & 1 & 31.44/0.870 & 30.10/\underline{0.863} & 30.29/\underline{0.853} & 30.86/0.855 & 29.54/0.843 & 29.79/0.838 \\
      &B2UB & 17 & 1 & \underline{31.64}/\underline{0.871} & \underline{30.25}/0.862 & \underline{30.46}/0.852 & \textbf{31.07}/\textbf{0.857} & \underline{29.92}/\underline{0.852} & \textbf{30.10}/\textbf{0.844} \\ & DCD-Net & 3 & 1 & \textbf{32.35}/\textbf{0.872} & \textbf{31.09}/\textbf{0.866} & \textbf{31.09}/\textbf{0.855} & \underline{31.00}/\textbf{0.857} & \textbf{29.99}/\textbf{0.855} & \underline{29.99}/\underline{0.843} \\
    \cmidrule(r){1-4} \cmidrule(lr){5-7} \cmidrule(lr){8-10} 
      \multirow{2}{*}{\shortstack[c]{\textit{Self-sup.} \\ \textit{(true BSN)}}} &Laine19 & 4 & 4 & 30.19/0.833 & 28.25/0.794 & 29.35/0.820 & 29.76/0.820 & 27.89/0.778 & 28.94/0.808 \\
      &SwinIA (ours)\textsuperscript{$\dag$} & 1 & 1 & 31.01/0.857 & 29.61/0.847 & 29.98/0.847 & 30.29/0.835 & 28.84/0.818 & 29.35/0.827 \\    
    \bottomrule
  \end{tabular}
  \caption{Denoising results on synthetic sRGB datasets along with the method description (supervision type and number of train and test passes). The highest PSNR(dB)/SSIM among self-supervised denoising methods is highlighted in \textbf{bold}, the second-best is \underline{underlined}. $^{\ast}$~denotes assuming known noise model. \textsuperscript{$\dag$}~denotes the models that we implemented and trained ourselves.}
  \label{tab:sRGB}
\end{table*}

\subsection{Mixture synthetic noise}
\label{sec:synth-mix}

We experiment with the sRGB natural images dataset (ImageNet) and grayscale Chinese characters dataset (H\`anZ\`i) with a mixture of multiple noise modalities, following~Xie~\etal~\cite{n2same}. ImageNet dataset was generated by randomly cropping 60\,000 patches of size $128 \times 128$ from the first 20\,000 images in ILSVRC2012~\cite{deng2009imagenet} validation set that consists of 50\,000 instances. We use 978 images for testing. Poisson noise ($\lambda = 30$), additive Gaussian noise ($\sigma = 60$), and Bernoulli noise ($p = 0.2$) were applied to the clean images before the training. 

H\`anZ\`i dataset consists of 78\,174 noisy images with 13\,029 different Chinese characters of size $64 \times 64$. Each noisy image is generated by applying Gaussian noise ($\sigma = 0.7$) and Bernoulli noise ($p = 0.5$) to a clean image. We select 10\% of images for testing and use the rest for training.

We present the results in \cref{tab:imagenet-hanzi}. SwinIA showed state-of-the-art performance on both datasets, outperforming not only Noise2Same and its recent modification Noise2Info but also Blind2Unblind. It also outperformed another BSN by Laine~\etal~\cite{laine2019high} by $+3.62$dB PSNR on average. 

\begin{figure*}[t!]
\centering
\includegraphics[width=\linewidth]{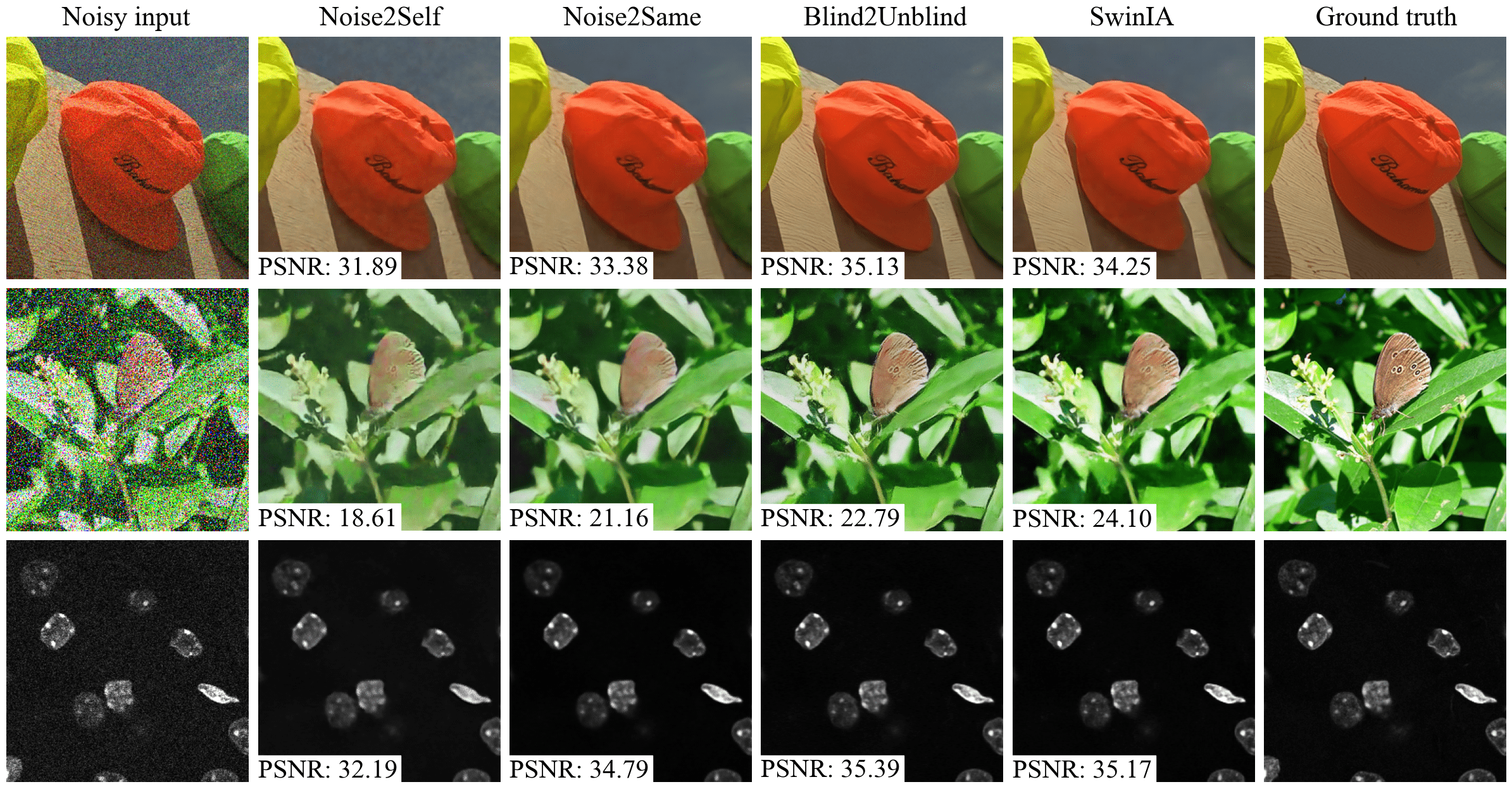}
\caption{Kodak~\cite{franzen1999kodak} (top), ImageNet~\cite{n2same} (middle), and FMD Two-Photon Mice~\cite{zhang2018poisson} (bottom) denoising examples. Every predicted image is cropped to a square for visualization presented with the corresponding PSNR score (in dB).}
\label{fig:viz-bsd-imagenet}
\end{figure*}

\subsection{Synthetic noise (sRGB)}
\label{sec:synth-rgb}
We follow~Huang~\etal~\cite{huang2021neighbor2neighbor} to create training and test sRGB datasets. For training, we select 44\,328 images between $256\times256$ and $512\times512$ pixels from the ILSVRC2012~\cite{deng2009imagenet} validation set. For testing, we use Kodak~\cite{franzen1999kodak}, BSD300~\cite{martin2001database}, and Set14~\cite{zeyde2010single}, repeated by 10, 3, and 20 times, respectively.
This adds up to 780 (240, 300, and 240) test images. We apply four types of noise in sRGB: Gaussian noise with (1) $\sigma=25$ and (2) $\sigma\in[5,50]$, Poisson noise with (3) $\lambda=30$ and (4) $\lambda\in[5,50]$.

We present the results in \cref{tab:sRGB}. SwinIA consistently supersedes the other BSN method by~Laine~\etal~\cite{laine2019high} and most of the mask-based methods, especially with Poisson noise, where we beat R2R by $0.5$dB PSNR on average. However, it did not compete with the state-of-the-art methods employing multiple passes in training and inference.

\subsection{Natural noise in fluorescent microscopy}
\label{sec:synth-micr}

We use Confocal Fish, Confocal Mice, and Two-Photon Mice datasets from the Fluorescent Microscopy Denoising Dataset~\cite{zhang2018poisson}. Each dataset consists of 20 views, each comprising 50 grayscale images of size 512$\times$512. Each image contains a \textbf{natural} mixture of Poisson and Gaussian noise. Ground truth is obtained by averaging all images in the view. We follow~Wang~\etal~\cite{wang2022blind2unblind} and select the 19th view for testing and the rest for training.  

The results are presented in \cref{tab:fmd}. SwinIA performed competitively across all datasets, yielding either best or second-best scores. Another BSN by Laine~\etal~\cite{laine2019high} required knowledge about the noise model and performed worse for the real-world data with noise mixture: SwinIA was better by $+0.91$dB PSNR and $+0.025$ SSIM on average for both Gaussian and Poisson assumed distributions. 

\begin{table}[t!]
\small
  \centering
  \begin{tabular}[b]{@{}lccc@{}}
    \toprule
      \multirow{2}{*}{Methods} &
      Confocal & Confocal & Two-Photon\\
      & Fish & Mice & Mice\\
    \midrule
      BM3D & 32.16/0.886 & 37.93/0.963 & 33.83/0.924 \\
    \midrule
      Noise2Clean & 32.79/0.905 & 38.40/0.966 & 34.02/0.925 \\
      Noise2Noise & 32.75/0.903 & 38.37/0.965 & 33.80/0.923 \\
    \midrule
      Laine19-pme (G) &  23.30/0.527 & 31.64/0.881 & 25.87/0.418 \\
      Laine19-pme (P) &  25.16/0.597 & 37.82/0.959 & 31.80/0.820 \\ 
    \midrule
      Noise2Void & 32.08/0.886 & 37.49/0.960 & 33.38/0.916 \\
      NBR2NBR & 32.11/0.890 & 37.07/0.960 & 33.40/\textbf{0.921} \\
      Noise2Self\textsuperscript{$\dag$} &  31.96/0.877 & 36.45/0.960 & 31.61/0.910 \\
      Noise2Same\textsuperscript{$\dag$} &  32.36/0.893 & 37.64/0.960 & 33.55/0.917 \\
      Blind2Unblind & \textbf{32.74}/\underline{0.897} & \textbf{38.44}/\underline{0.964} & \textbf{34.03}/0.916 \\
      CADT & 32.52/0.895 & \underline{38.21}/0.962 & 33.64/0.914 \\
      \midrule
      Laine19 (G) & 31.62/0.849 & 37.54/0.959 & 32.91/0.903 \\
      Laine19 (P) & 31.59/0.854 & 37.30/0.956 & 33.09/0.907 \\ 
      SwinIA (ours)\textsuperscript{$\dag$} & \underline{32.65}/\textbf{0.904} & \underline{38.21}/\textbf{0.966} & \underline{33.90}/\underline{0.920} \\ 
    \bottomrule
  \end{tabular}
  \caption{Denoising results on Fluorescent Microscopy datasets. The highest PSNR(dB)/SSIM among self-supervised denoising methods is highlighted in \textbf{bold}, while the second-best is \underline{underlined}. For Laine~\etal~\cite{laine2019high}, G --- Gaussian, P --- Poisson. \textsuperscript{$\dag$} denotes the models that we implemented and trained ourselves.}
  \label{tab:fmd}
\end{table}

\subsection{Ablation study}
\label{sec:ablation}

We ran ablation experiments on synthetic noise grayscale datasets to validate our key architecture design elements. The results are summarized in \cref{tab:ablation}. In particular, we experimented with alternative architectures without pixel shuffle: dilated attention~\cite{hassani2022dilated} and flat architecture without encoder-decoder separation~\cite{swinir}. Both not only harmed the performance but also considerably increased training time. Removing the mask on inference led to a valuable performance increase, which proves the necessity of unblinding. 

We separately tested attention masking and input isolation as unavoidable restrictions for self-unawareness, and the full encoder flow~--- a configuration where the input is sequentially propagated down the encoder blocks allowing context awareness inside of patches. All three experiments resulted in learning the identity function and poor scores.

\begin{table}[t!]
\small
  \centering
  \begin{tabular}[b]{lccc}
    \toprule
       \multirow{2}{*}{Experiment} & 1 epoch & \multirow{2}{*}{BSD68} & \multirow{2}{*}{Set12} \\
        & (min) & & \\
    \midrule
       Our best & 11.5 & \textbf{30.01}/\textbf{0.837} & 29.61/\textbf{0.848} \\
    \midrule
       Dilated attention & 14.5 & \underline{29.90}/\underline{0.830}  & \textbf{29.63}/\underline{0.842} \\
       Flat architecture & 14.5 & 29.87/0.826 & \underline{29.62}/0.840 \\
       Masked inference & 11.5 & 29.35/0.816 & 28.92/0.832 \\
       No attention mask\textsuperscript{$\ast$} & 11.5 & 20.45/0.380 & 20.33/0.394 \\
       No input isolation\textsuperscript{$\ast$} & 12 & 20.51/0.379 & 20.39/0.383  \\
       Full encoder\textsuperscript{$\ast$} & 11.5 & 21.22/0.396 & 21.01/0.402  \\
       \midrule
       Larger window (12) & 47 & 30.09/0.840 & 29.69/0.850 \\
       Smaller window (6) & 7 & 29.79/0.830 & 29.37/0.842 \\
    \bottomrule
  \end{tabular}
  \caption{Ablation results on grayscale data with synthetic Gaussian noise $(\sigma = 25)$. The experiments marked with \textsuperscript{$\ast$} ended up learning the identity function.}
  \label{tab:ablation}
\end{table}

We also experimented with the attention window size. Larger window size $w$ requires larger training crops $p$ because of downsampling and is computationally expensive since the attention computation complexity is $\Theta(w^4)$. For $w=12,p=96$, the training was four times longer. The increased context provided a marginal gain of +0.08/+0.003 PSNR/SSIM on average, while decreasing it to $w=6,p=48$ reduced the scores by -0.23/-0.006 (see \cref{tab:ablation}).

\section{Discussion}
\label{sec:discussion}

The flexibility of transformer architecture allowed us to build an assumption-free SwinIA model that has many strengths. First, it is robust across various noise types and image modalities. Most notably, it achieves state-of-the-art performance for the most complex synthetic mixed noise datasets and several others. Second, our model is optimized by minimizing a simple loss function without tunable hyperparameters. The main competitors, Blind2Unblind~\cite{wang2022blind2unblind} and DCD-Net~\cite{Zou_2023_ICCV}, have multiple empirically set loss constants changing according to the selected training schedule. Finally, SwinIA uses a single forward pass in both training and inference allowing to decrease time and compute, which is especially important for a transformer-based model (most competitor models use multiple passes, as we report in our tables with results). In our experiments, SwinIA trains twice faster than Blind2Unblind on the same hardware (see~\cref{sec:training} for the details).

The versatility of a true blind-spot model comes with limitations. The pixel itself contains the most useful information about its true signal, which is inevitably lost in the training process. However, we are able to remove the attention mask during inference and allow pixels to attend to their initial values. This is not possible in a convolutional BSN where hiding is done through zeroing trainable weights~\cite{honzatko2020efficient,wu2020unpaired} or excluding the central pixel from the field of view~\cite{laine2019high}. We further discuss and visualize our unmasking in~\cref{sec:attention}. Also, a BSN assumes spatially uncorrelated noise, which is not the case for many digital photography datasets because of hardware pixel interpolation. This problem can be mitigated with increased patch size or dilated attention resembling the approach by Lee~\etal~\cite{lee2022ap}.  

\begin{figure}[t]
\centering
\includegraphics[width=\linewidth]{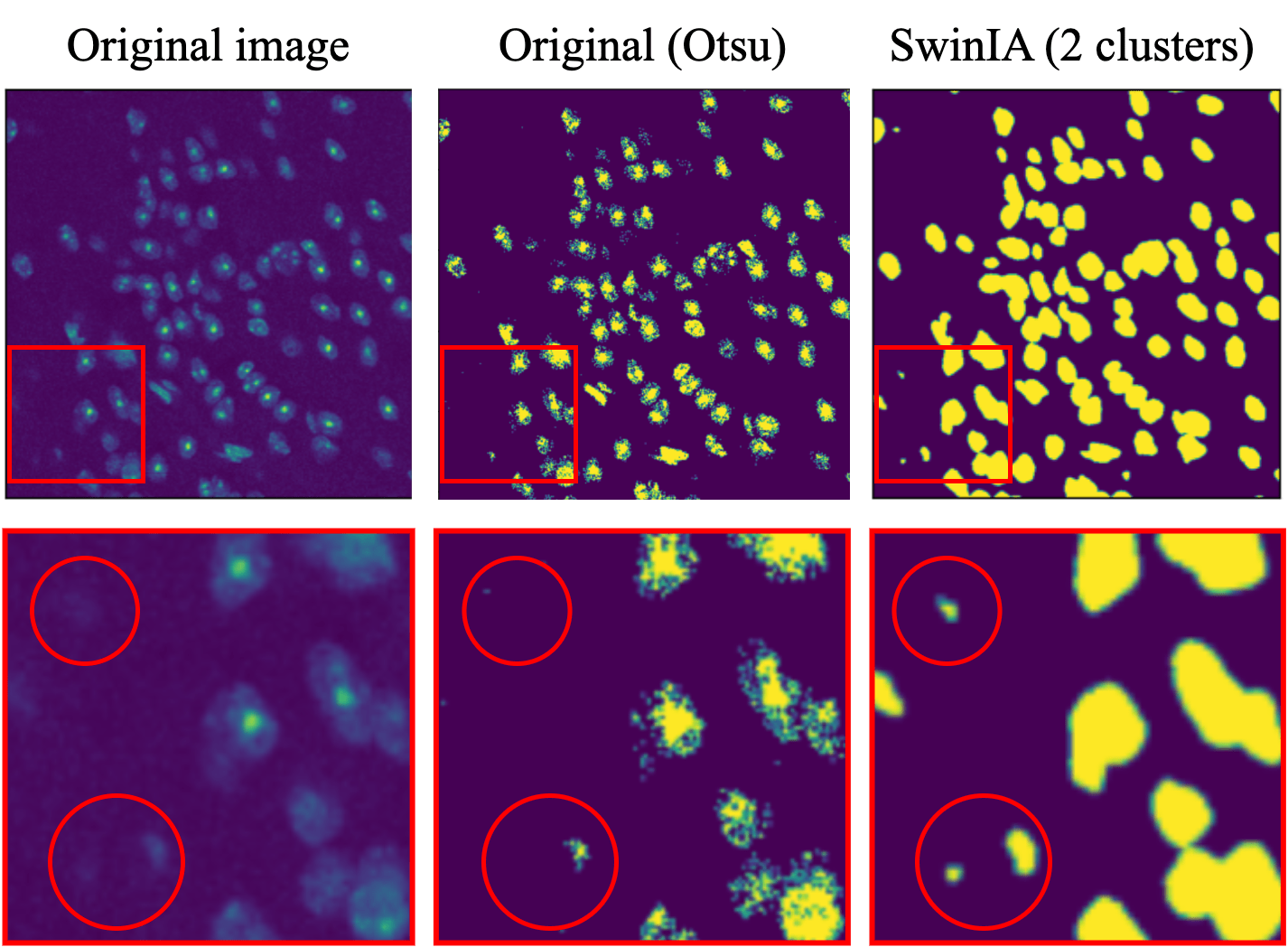}
\caption{Binary thresholding on FMD Confocal Mice. We apply Otsu thresholding~\cite{otsu} to the original image (middle column) and $k$-means with $k=2$ to the final feature map of SwinIA (right column). A part of each image is zoomed (bottom row), and the blurry and half-light cells are highlighted with red circles.}
\label{fig:seg}
\end{figure}

Transformers are known for their ability to extract rich representations from large datasets, and we expect our method to improve with increasing training set size. Besides, being conceptually similar to the language modeling objective~\cite{shin-etal-2020-fast}, our solution could be used in self-supervised pre-training to produce \textbf{pixel} embeddings for downstream image tasks. \cref{fig:seg} shows an example of SwinIA embeddings clustering into segmentation masks. \cref{fig:seg} also features Otsu thresholding~\cite{otsu} to ensure that quality masks are not simply obtainable straight from the noisy input. In~\cref{sec:embedding}, we show more examples of feature clustering, also comparing to other models. We leave further investigation of the feature extraction abilities for future work.

\section{Conclusion}

We propose SwinIA, the first convolution-free transformer architecture for blind-spot self-supervised denoising. Unlike its counterparts, it does not require access to clean data or assume any noise distribution. SwinIA also does not use input masking and can be trained in an autoencoder fashion with a single forward pass and an MSE loss. Finally, it does not require multivariate hyperparameter tuning and achieves competitive results, outperforming state-of-the-art methods on several common benchmarks and showing robustness to different kinds of synthetic and natural noise in images of various modalities.
\section*{Acknowledgements}

This work was supported by Revvity\footnote{\url{https://www.revvity.com/}}. The authors acknowledge the significant computational resources provided by the High-Performance Computing Cluster of the University of Tartu~\cite{hpc}. The authors also express high gratitude to their colleagues from the Institute of Computer Science, University of Tartu, particularly the Biomedical Computer Vision research group members and Elizaveta Korotkova, for their continuous help and support. 

{\small
\bibliographystyle{ieee_fullname}
\bibliography{paper}
}

\begin{figure*}[t!]
  \centering
  \includegraphics[width=\linewidth]{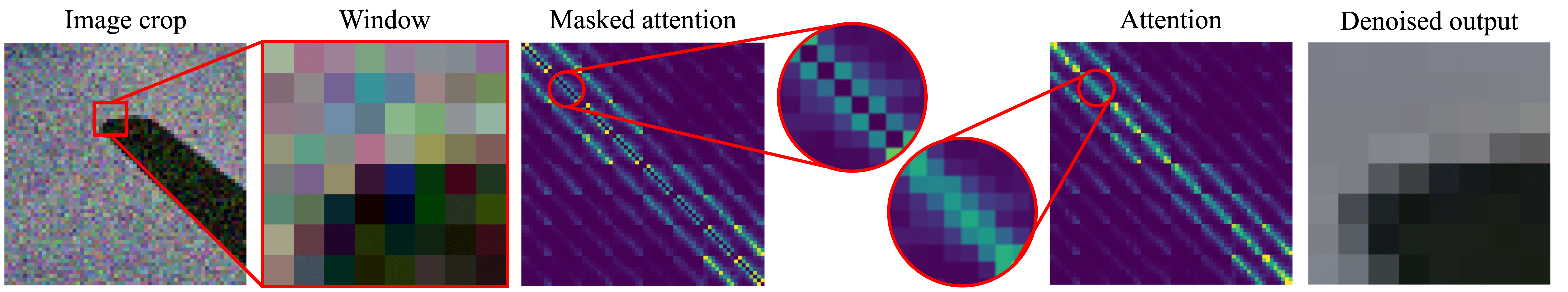}
   \caption{Example of masked (training, \cref{eq:msa}) and unmasked (inference, \cref{eq:msai}) window attention on an $8\times 8$ window. The window contains $64$ pixels, thus the maps are of size $64\times64$. The mask is applied additively before the softmax operation (\cref{sec:transformer}).}
   \label{fig:window}
\end{figure*}

\newpage

\appendix

\section{Attention mask justification}
\label{sec:attention}

Opposed to the multitude of methods based on pixel masking, our approach does not utilize random masks. More importantly, we do not mask pixels at all. Pixel mask, unlike our attention mask, has many disadvantages: it implies tuning hyperparameters, increased training time for multiple forward passes, or slower convergence. Furthermore, masked pixels may appear in the receptive field of their neighbors in training, thus affecting the denoising.  

The attention is a parameter-free matrix, a dot product between trained embedding projections. Therefore, our attention mask neither hides any trainable parameters nor alters the input but simply cancels attention of each pixel to itself during training ($1.6\%$ of the full matrix). It enables our model to learn meaningful embeddings through neighboring pixel interaction without collapsing to identity. 
During inference, the attention scores on diagonal will remain high despite the masking during training, because they reflect pixel self-similarity. Therefore, it is only natural to unhide the main diagonal to let the model propagate the signal directly from each pixel. Fig.~\ref{fig:window} illustrates this unblinding by comparing attention maps.

\begin{figure}[t]
\centering
\includegraphics[width=\linewidth]{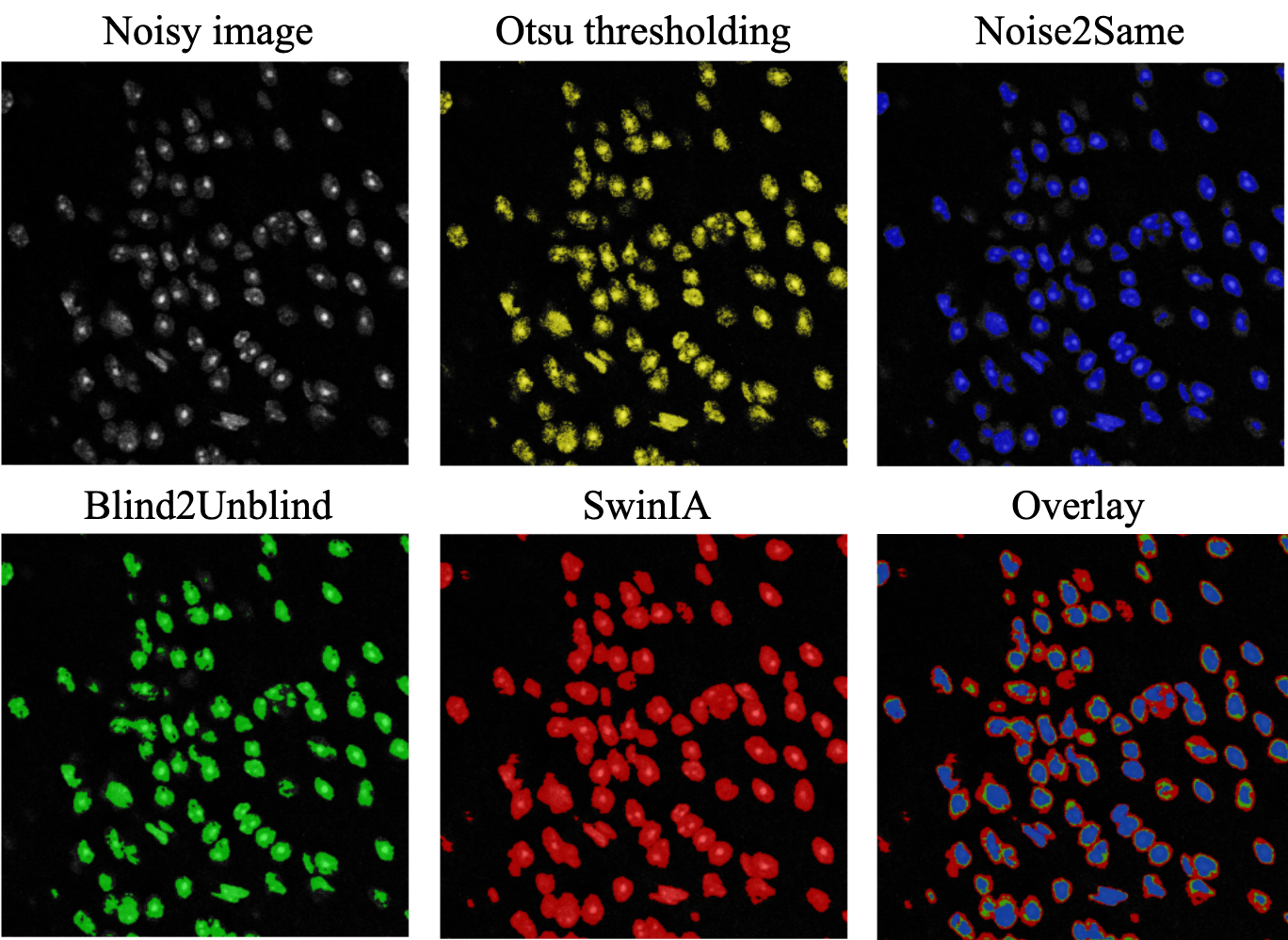}
\caption{Binary masks produced by Otsu~\cite{otsu} and $k$-means clustering on model feature maps on FMD Confocal Mice~\cite{zhang2018poisson}.}
\label{fig:overlay-cf}
\end{figure}

\begin{figure}[t]
\centering
\includegraphics[width=\linewidth]{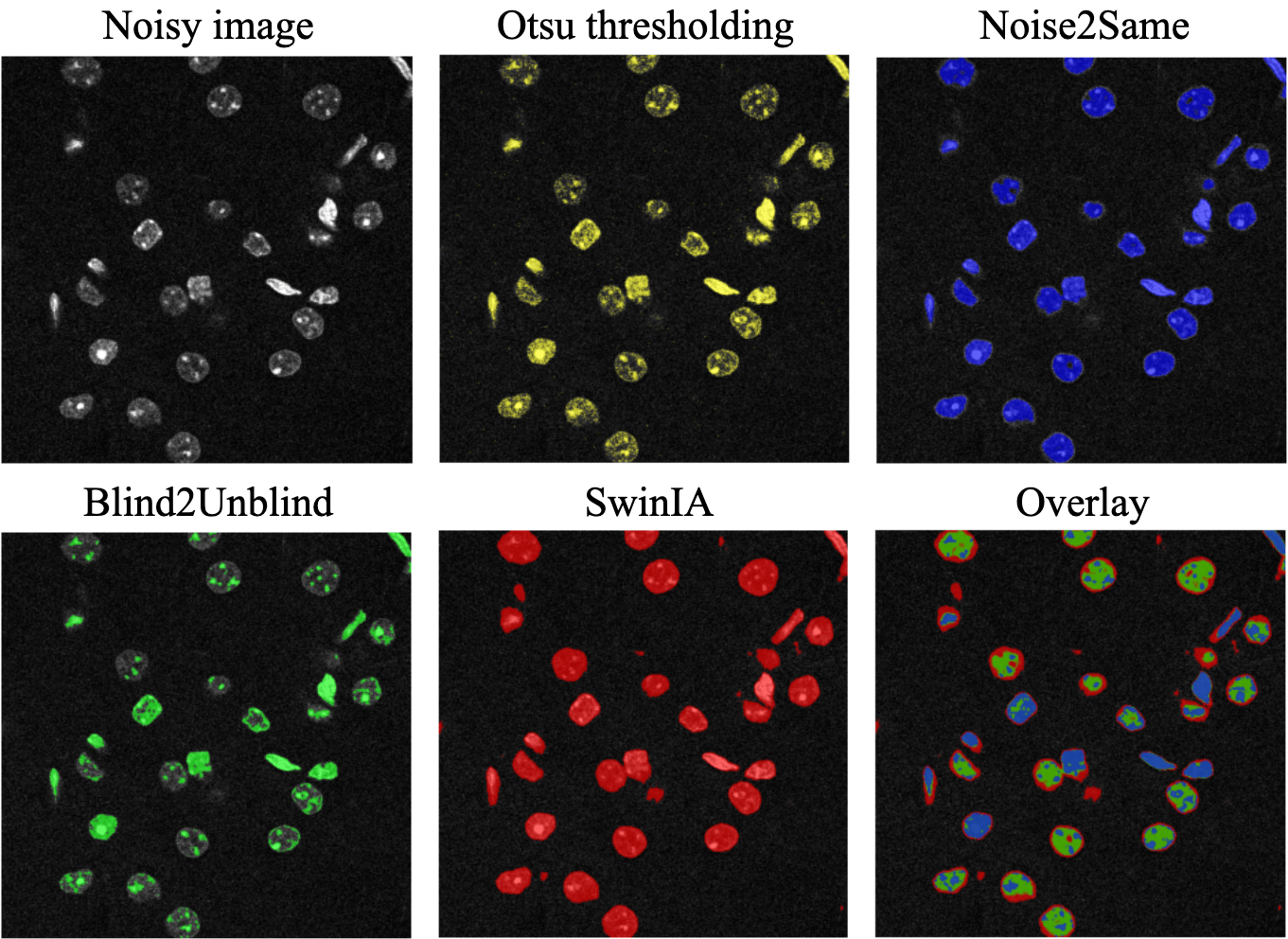}
\caption{Binary masks produced by Otsu~\cite{otsu} and $k$-means clustering on model feature maps on FMD Two-Photon Mice~\cite{zhang2018poisson}.}
\label{fig:overlay-tp}
\end{figure}

\begin{table*}[t!]
  \centering
  \begin{tabular}[b]{lcccccc}
    \toprule
      \multirow{2}{*}{Dataset} & \multicolumn{2}{c}{Noise2Same} & \multicolumn{2}{c}{Blind2Unblind} & \multicolumn{2}{c}{SwinIA} \\
       & TT (h) & AIT (ms) & TT (h) & AIT (ms) & TT (h) & AIT (ms) \\
    \midrule
      Synthetic (sRGB) & 4 & 26 &---&35& 10 & 416  \\
      Synthetic (grayscale) & 2 & 12 &---&---& 10 & 239  \\
      ImageNet & 1 & 14 &25&206& 10 & 554  \\
      H\`anZ\`i & 1 & 5 &6&6& 10 & 29  \\
      Microscopy & 1.5 & 20 &---&32& 4 & 415  \\
    \bottomrule
  \end{tabular}
  \caption{Comparison of training time (TT) in hours, and average inference time (AIT) on the test set in milliseconds of Noise2Same~\cite{n2same}, Blind2Unblind~\cite{wang2022blind2unblind}, and SwinIA~(ours) on various datasets. For Blind2Unblind~\cite{wang2022blind2unblind}, we report the results of the experiments that we ran or re-evaluated ourselves, the code and the weights for experiments with synthetic grayscale noise are missing from the official repository.}
  \label{tab:time}
\end{table*}
\begin{table*}[t!]
  \centering
  \begin{tabular}[m]{lcccc}
    \toprule
      Criterion & Noise2Same & Blind2Unblind & SwinIR & SwinIA \\
    \midrule
      Number of trainable parameters & 5.564M & 1.100M & 4.610M & 3.966M  \\
      FLOPs/image $1\times64\times64$ (training) & 10.002G & 19.639G & 18.978G &  15.890G   \\
      FLOPs/image $1\times64\times64$ (inference) & 5.001G & 1.155G & 18.978G  & 15.890G   \\
    \bottomrule
  \end{tabular}
  \caption{Comparison of the number of parameters and FLOPS between Noise2Same~\cite{n2same}, Blind2Unblind~\cite{wang2022blind2unblind}, SwinIR~\cite{swinir}, and SwinIA~(ours). The number of FLOPs is calculated separately for training and inference.}
  \label{tab:flops}
\end{table*}

\section{Embeddings analysis}
\label{sec:embedding}

We further investigate the features extracted by SwinIA in the last transformer block. In \cref{fig:overlay-cf,fig:overlay-tp}, we continue the comparison of binary segmentations performed on the original image with Otsu thresholding~\cite{otsu} and on the final feature maps of the models with $k$-means clustering into two clusters. We also include the clusterings of feature maps produced by Noise2Same~\cite{n2same} and Blind2Unblind~\cite{wang2022blind2unblind}. For better comparison, we include the overlay of model embedding clusterings in these figures.

Compared over two microscopy datasets, clusters on the embeddings produced by SwinIA are more complete, sharp, and close to the shape of the cells than those of the counterparts. SwinIA also follows image semantics rather than raw pixel values in its embeddings: on Two-Photon Mice, every highlighted object is a cell, while for Blind2Unblind~\cite{wang2022blind2unblind} clusters represent either full objects or their subcomponents depending on their brightness (note that the result is similar to Otsu intensity thresholding). It is worth noting that there is a high similarity between the embeddings of the pixels that are far from each other and never participate in the same attention window. This confirms that the training of SwinIA is well-regularized and aimed at extracting meaningful image features, equally good in global and local contexts. Embedding clusters show our model's potential as a universal self-supervised feature extractor for dense downstream tasks such as semantic segmentation.

\section{Training details}
\label{sec:training}

We train SwinIA for 50\,000 steps with batch size 64 and use Lion~\cite{chen2023symbolic} with one cycle schedule~\cite{Smith2018SuperconvergenceVF} warming up for 15\% of steps and then reducing learning rate from $3\times10^{-4}$ to $10^{-6}$. In experiments with FMD, we decrease the number of steps to 20\,000 and the peak learning rate to $10^{-4}$, because the dataset is small. During training, we randomly cut $64 \times 64$ crops from images scaled to $[0, 1]$, rotate them by multiples of $90^{\circ}$, and flip them horizontally and vertically. Each crop is standardized with $\mu$ and $\sigma$ of the training dataset. During validation, we pad each image for divisibility by 32 with reflection and crop the padding after prediction. For ImageNet in the mixed noise experiment (\cref{sec:synth-mix}), we denoised overlapping tiles for several largest images and stitched them back before evaluation.

We implemented all models in Python 3.8.3 and PyTorch 1.12.1~\cite{paszke2017automatic} and trained them on NVIDIA A100 80GB GPUs (driver version: 470.57.02, CUDA version: 11.4). We used \texttt{einops}~\cite{einops} for tensor permutations.

We compare the training time and average inference time of Noise2Same~\cite{n2same}, Blind2Unblind~\cite{wang2022blind2unblind}, and SwinIA models in \cref{tab:time} and show the difference in the number of parameters and the number of floating point operations (FLOPs) per grayscale image in \cref{tab:flops}.

\section{Denoising results visualization}

\begin{figure*}[ht]
\centering
\setlength{\abovecaptionskip}{0.1cm} 
\includegraphics[width=\linewidth]{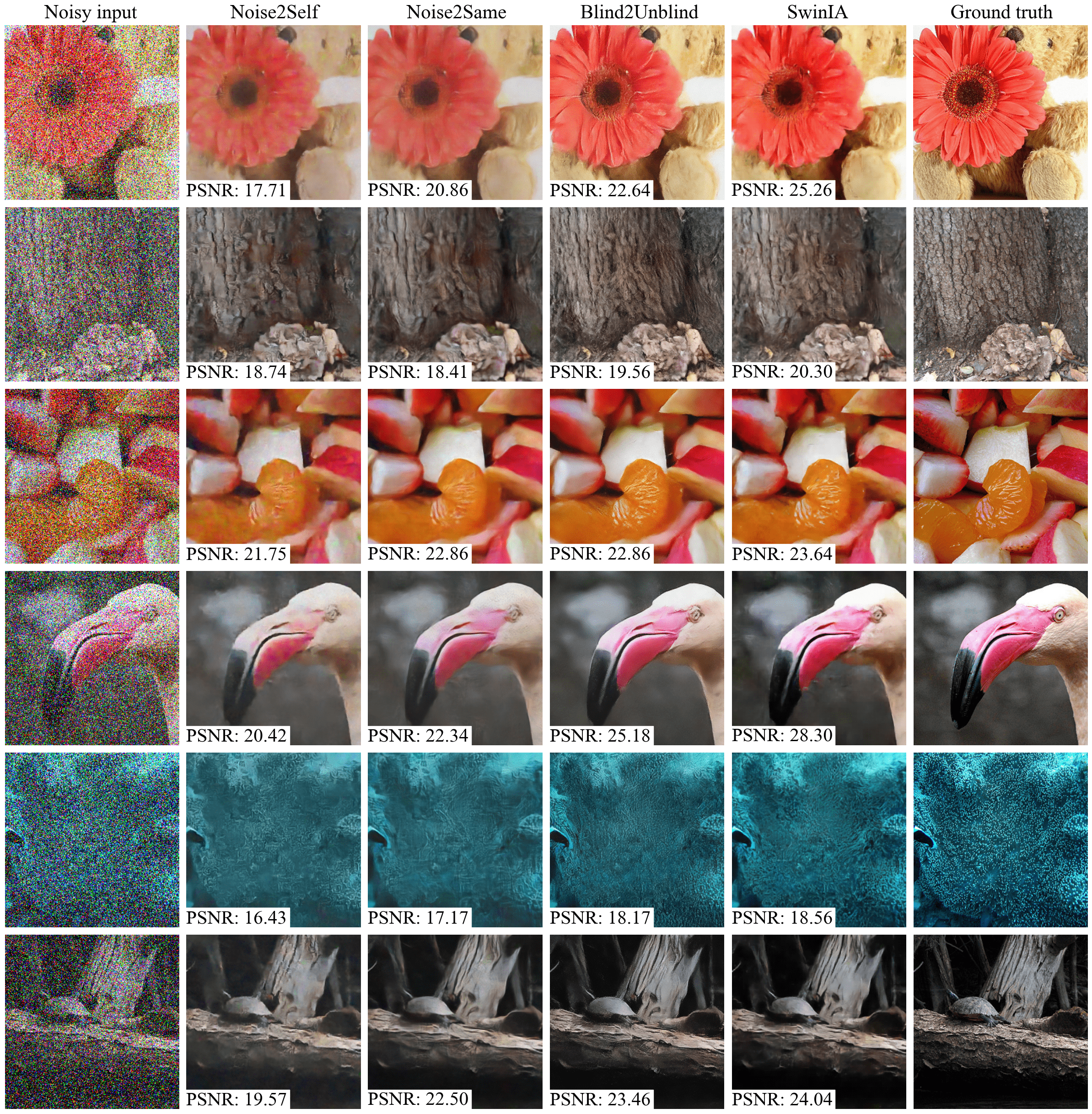}
\caption{Denoising examples on ImageNet dataset with mixed synthetic noise~\cite{n2same}. Each row contains noisy and ground truth images, along with the predictions of Noise2Self~\cite{n2self}, Noise2Same~\cite{n2same}, Blind2Unblind~\cite{wang2022blind2unblind}, and SwinIA~(ours) models with corresponding PSNR scores. Each image is center-cropped for visualization.
}
\label{fig:supp-imagenet}
\end{figure*}

\begin{figure*}[ht]
\centering
\setlength{\abovecaptionskip}{0.1cm} 
\includegraphics[width=\linewidth]{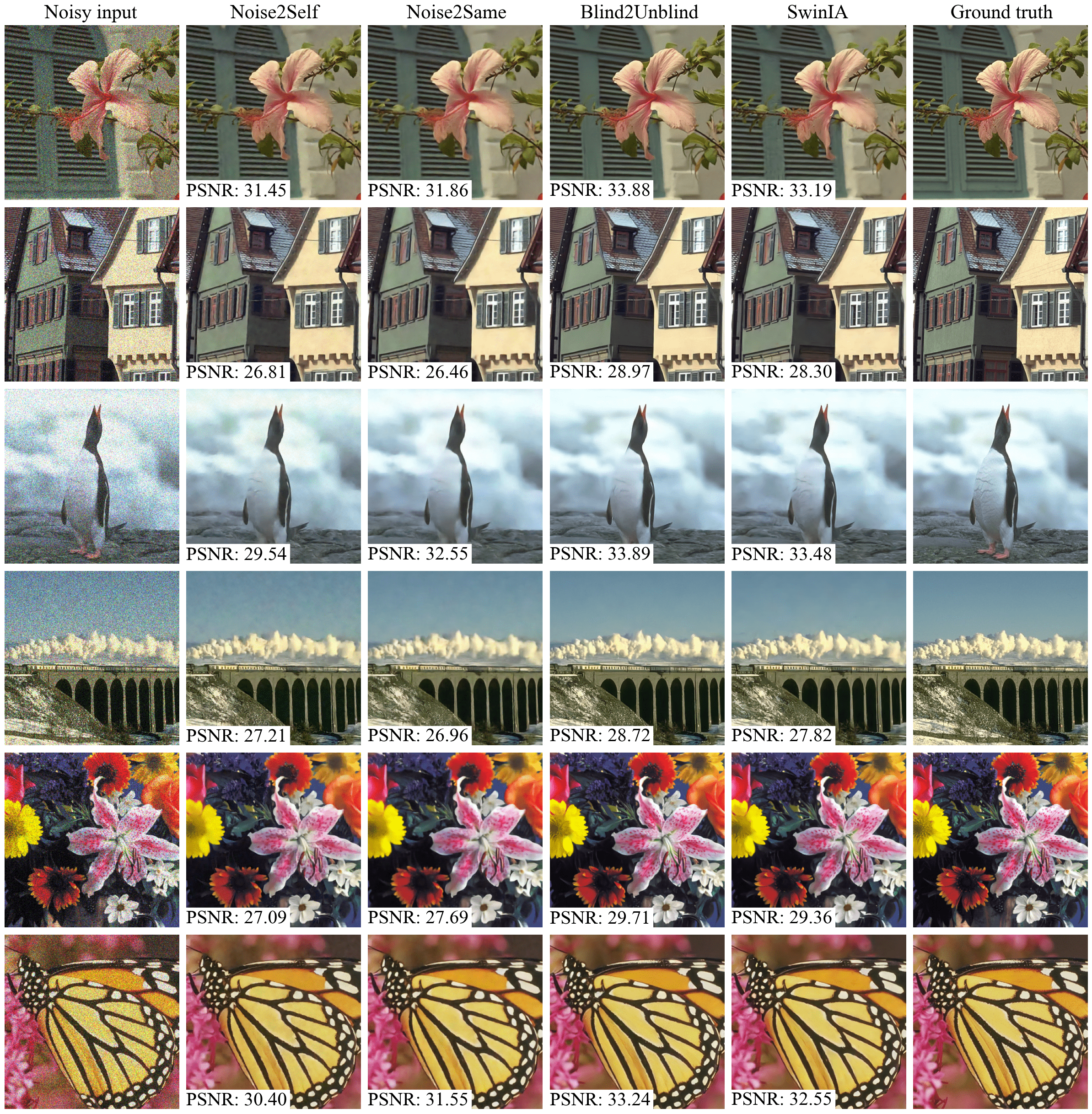}
\caption{Denoising examples on sRGB data with synthetic Poisson noise ($\lambda=30$). Each row contains noisy and ground truth images, along with the predictions of Noise2Self~\cite{n2self}, Noise2Same~\cite{n2same}, Blind2Unblind~\cite{wang2022blind2unblind}, and SwinIA~(ours) models with corresponding PSNR scores. Each image is center-cropped for visualization.}
\label{fig:supp-synthetic}
\end{figure*}

\begin{figure*}[ht]
\centering
\setlength{\abovecaptionskip}{0.1cm} 
\includegraphics[width=\linewidth]{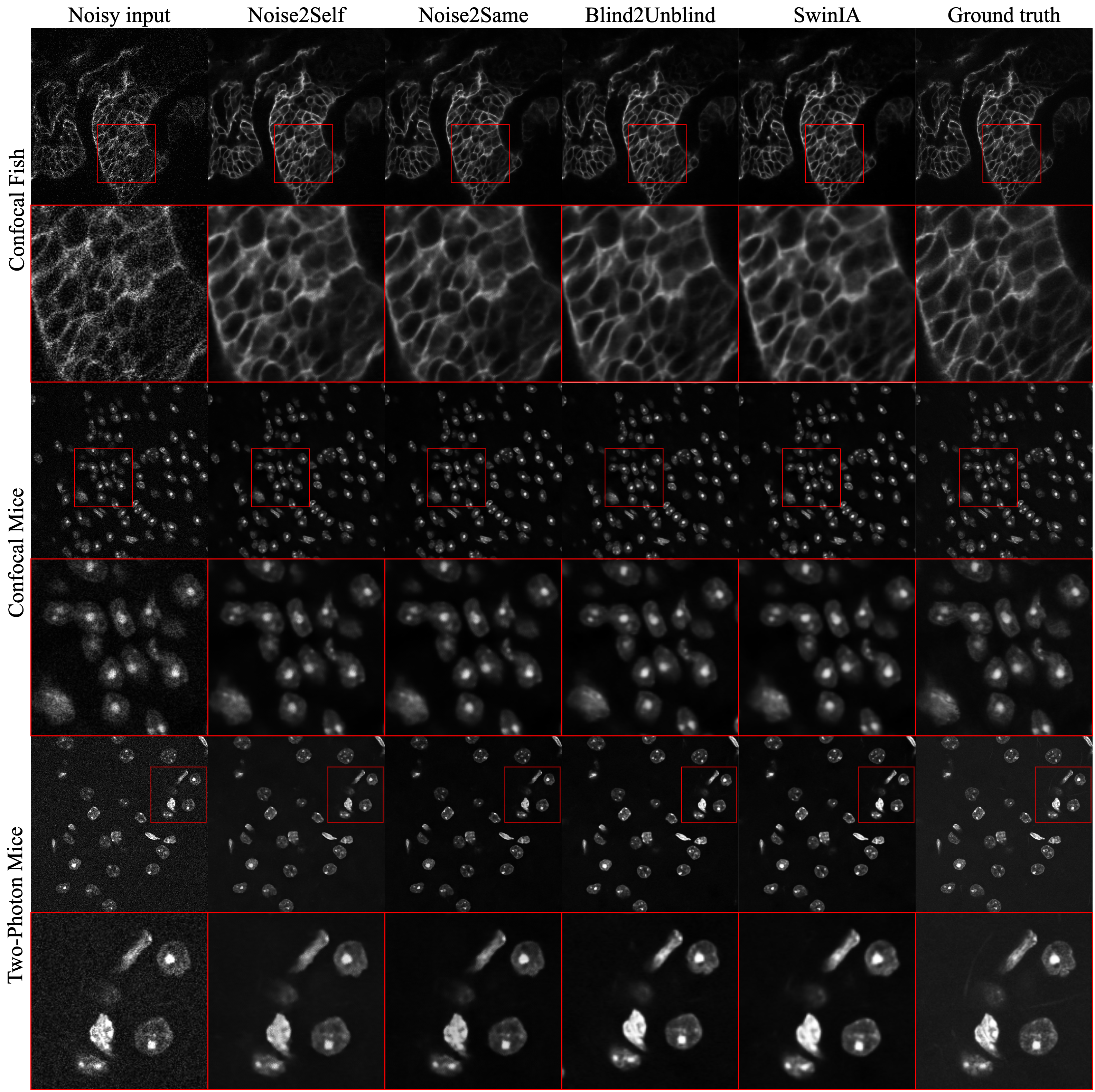}
\caption{Denoising examples on fluorescent microscopy images with natural noise~\cite{zhang2018poisson}. Each pair of rows contains noisy and ground truth images from the three used datasets (top~--- full size, bottom~--- zoomed in for a better view), along with the predictions of Noise2Self~\cite{n2self}, Noise2Same~\cite{n2same}, Blind2Unblind~\cite{wang2022blind2unblind}, and SwinIA~(ours). }
\label{fig:supp-fmdd}
\end{figure*}

\noindent In \cref{fig:supp-imagenet,fig:supp-synthetic,fig:supp-fmdd}, we show additional denoising examples in experiments with synthetic Poisson noise, mixed synthetic noise, and real-world noise on microscopy data, respectively. 

\section{Model card}
We present a model card with the main information and technical details about our SwinIA model in \cref{tab:model_card}.

\FloatBarrier
\onecolumn
\begin{longtable}[c]{ p{.15\textwidth}  p{.65\textwidth} }
\caption{Model Card of SwinIA.}
\label{tab:model_card}
\endfirsthead
\endhead
\toprule
\multicolumn{2}{c}{\textbf{Model Summary}}      \\ \midrule
\multicolumn{1}{l}{Model Architecture} & 
Fully transformer-based image autoencoder model for end-to-end self-supervised image denoising with no convolutions. For details, see~\cref{sec:methods}.  \\ 
\midrule
\multicolumn{1}{l}{Input(s)} & 
The model takes noisy images as input, batch and channel dimensions go first. %
\\ \midrule
\multicolumn{1}{l}{Output(s)} & 
The model outputs a batch of denoised images of the same shape as input. %
\\
\midrule
\multicolumn{2}{c}{\textbf{Usage}}      \\ \midrule
\multicolumn{1}{l}{Application} & 
The model can be used in self-supervised image denoising for any type of spatially-uncorrelated synthetic and natural noise on both grayscale and colored images. Also, it is theoretically possible to use the model in self-supervised pre-training for extracting features from images for downstream vision tasks.
\\ 
\midrule
\multicolumn{1}{l}{Known Limitations} & 
The model is computationally expensive and requires both powerful GPU hardware and considerable training time. Also, small datasets will most probably lead to overfitting. Finally, the model will most likely learn an identity function on the data with spatially correlated noise without proper tuning of patch sizes (as mentioned in~\cref{sec:discussion}).
\\
\midrule

\multicolumn{2}{c}{\textbf{System Type}}      \\ \midrule
\multicolumn{1}{l}{System Description} & This is a standalone model.  \\ \midrule
\multicolumn{1}{l}{Dependencies} & None.  \\ 
\midrule
\multicolumn{2}{c}{\textbf{Implementation Frameworks}}      \\ \midrule
\multicolumn{1}{l}{Hardware \& Software} & 
\begin{minipage}{0.65\textwidth}
Hardware: NVIDIA A100 80GB GPUs (driver version: 470.57.02, CUDA version: 11.4). 
\\\\
Software: Python 3.8.3, PyTorch 1.12.1~\cite{paszke2017automatic}, einops~\cite{einops}.
\end{minipage}

\\ \midrule

\multicolumn{1}{l}{Compute Requirements} & In every experiment, SwinIA was trained on one NVIDIA A100 80GB GPU for different numbers of steps (see~\cref{sec:results} for details).
\\
\midrule
\multicolumn{2}{c}{\textbf{Model Characteristics}}      \\ \midrule
\multicolumn{1}{l}{Model Initialization} & The model is trained from a random initialization.  \\ \midrule
\multicolumn{1}{l}{Model Status} & This is a static model trained on offline datasets.  \\ \midrule
\multicolumn{1}{l}{Model Stats} & SwinIA model has 3.966 million trainable parameters and performs 15.89 GFLOPS (floating point operations per second) per $64\times64$ grayscale image. \\
\midrule
\multicolumn{2}{c}{\textbf{Data Overview}}      \\ \midrule
\multicolumn{1}{l}{Training Datasets} & \begin{minipage}{0.65\textwidth}
Synthetic noise (sRGB): ILSVRC2012~\cite{deng2009imagenet} validation set. 
\\ \\
Synthetic noise (grayscale): BSD400~\cite{zhang2017beyond}.
\\ \\
Mixture synthetic noise: ImageNet~\cite{n2same}, H\`anZ\`i~\cite{n2self,n2same}.
\\ \\
Natural noise (grayscale): Confocal Fish, Confocal Mice, and Two-Photon Mice datasets from the Fluorescent Microscopy Denoising Dataset~\cite{zhang2018poisson}.

\end{minipage} \\
\midrule
\pagebreak
\midrule
\multicolumn{1}{l}{Evaluation Datasets} & \begin{minipage}{0.65\textwidth}
Synthetic noise (sRGB): Kodak~\cite{franzen1999kodak}, BSD300~\cite{martin2001database}, Set14~\cite{zeyde2010single}. \\ \\
Synthetic noise (grayscale): Set12 and BSD68~\cite{roth2005fields}.
\\ \\
Mixture synthetic noise: ImageNet~\cite{n2same}, H\`anZ\`i~\cite{n2self,n2same}.
\\ \\
Natural noise (grayscale): Confocal Fish, Confocal Mice, and Two-Photon Mice datasets from the Fluorescent Microscopy Denoising Dataset~\cite{zhang2018poisson}.
\end{minipage} \\
\bottomrule
\end{longtable}

\end{document}